\documentclass[11pt]{article}

\usepackage[preprint]{acl}

\usepackage{times}
\usepackage{latexsym}

\usepackage[T1]{fontenc}
\usepackage{array}
\usepackage[utf8]{inputenc}

\usepackage{microtype}

\usepackage{inconsolata}

\usepackage{graphicx}
\usepackage{xcolor}
\usepackage{booktabs}
\usepackage{amsmath}
\usepackage{fancyvrb}
\title{Learning from Unreliable Operator References: Claim-Level Rewards for Customer Support Agents}

\author{
  \textbf{Dmitrii Stoianov},
  \textbf{Olga Tsymboi},
  \textbf{Danil Taranets},
  \textbf{Daniil Strizhakov},
  \textbf{Ilya Buryak},
\\
  \textbf{Almaz Dautov},
  \textbf{Dmitry Fatnev},
  \textbf{Arthur Babkin},
  \textbf{Aleksandr Fida},
  \textbf{Daniil Dryabin},
\\
  \textbf{Maksim Shilenko},
  \textbf{Nikita Surnachev},
  \textbf{Anatolii Potapov}
\\
\\
  T-Tech
\\
  \small{
    \textbf{Correspondence:} \href{mailto:anatolii.s.potapov@gmail.com}{anatolii.s.potapov@gmail.com}
  }
}

\begin{document}
\maketitle

\begin{abstract}
Even a correct operator reply can be an unreliable training target
for a customer-support agent. About a quarter of the factual claims in these replies lacked full support in the agent's knowledge base and tool records. We treat operator replies as claims to verify rather than answers to imitate. Unverified claims guide revisions to the knowledge base and tools. Verified claims serve as coverage targets in a claim-level GRPO reward that scores factual precision against the evidence the agent retrieved for its answer. On an offline benchmark, 8.6\% of the GRPO agent's text answers were flagged for hallucinations, compared with 13.6\% for the supervised fine-tuning baseline. In a randomized A/B test against the existing procedural bot, the deployed system increased the share of conversations closed without a human operator from 63.3\% to 80.8\%.
\end{abstract}

\section{Introduction}
\label{sec:intro}

Operator replies are unreliable references for training customer-support agents because even accurate answers can draw on records the agent cannot access. Historical replies are valuable because
they show how operators answer recurring questions
\citep{laskar-etal-2025-ai}, while the surrounding conversations
reveal the steps they take to resolve them \citep{choubey2025turning}.
Deployed agents already learn service strategies from expert dialogues
\citep{gao2026benchmarking} and next actions from operator traces
\citep{borovkov2026learning}. The risk arises when an operator bases
a reply on customer records the agent cannot access. As in other
settings with better-informed demonstrators
\citep{weihs2021bridging,pmlr-v139-warrington21a}, direct imitation
can then teach the agent to state a fact it cannot verify.

\begin{figure*}[t]
\centering
\includegraphics[width=\textwidth]{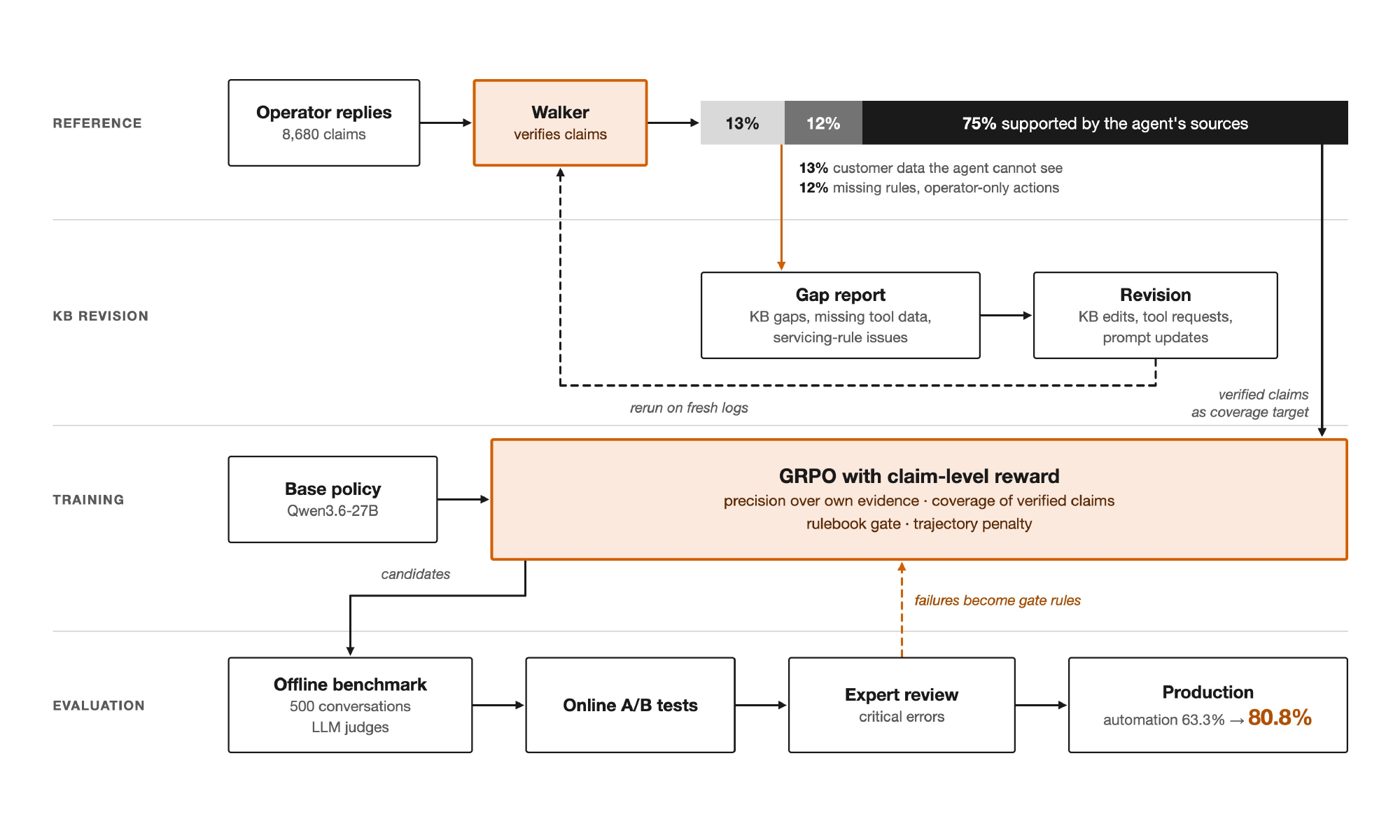}
\caption{Overview: unverified claims drive knowledge-base
and tool revision; verified claims serve as the coverage target of a
claim-level GRPO reward; candidates pass an offline benchmark, online
A/B tests, and expert review before deployment, and review failures
become gate rules. }
\label{fig:pipeline}
\end{figure*}

In our referral program's support logs, about a quarter of the factual
claims in operator replies could not be fully verified against the
agent's knowledge base and tool records. Narrowing this gap requires
sustained work to document missing rules, keep them current, and
adapt legacy tools for secure access to customer data. Some records
also remain restricted to human operators. How can we learn from
operator replies given the evidence the deployed agent can actually
access?

We verify operator replies one factual claim at a time, as in
claim-level factuality evaluation \citep{min2023factscore}. Claims
we cannot verify help identify gaps in the knowledge base or tools.
Verified claims define coverage targets for group relative policy
optimization (GRPO) \citep{grpocite}. The reward scores coverage
against these targets and factual precision against the evidence
the agent actually retrieved when producing its answer
(Figure~\ref{fig:pipeline}).

The resulting GRPO agent had an offline hallucination rate
5 p.p. lower than that of the same base model trained
on filtered teacher trajectories with supervised fine-tuning (SFT). In a randomized A/B test with about
17500 conversations per arm, the agent increased the share of
conversations closed without a human operator by 17.5 p.p. and reduced operator time per customer by 30.4\% relative
to the procedural bot. The agent now handles live referral-support
requests.

Our main contributions are:
\begin{itemize}
    \item A verification loop that identifies gaps in the
    knowledge base and tools while extracting supported
    operator claims for training.
    \item A claim-level RL formulation for learning from verified
    operator claims that jointly optimizes grounding,
    answer coverage, and compliance with operational
    requirements.
    \item Empirical insights into how supervision, retrieval,
    and reward design shape agent reliability and automation,
    including the finding that higher offline helpfulness scores did not translate into higher
automation.
\end{itemize}
 
\section{Related Work}
\label{sec:related}

\paragraph{Claim-level factuality and rewards.}
FActScore evaluates factual precision by decomposing generated text into 
atomic claims \citep{min2023factscore}. RAGAS assesses faithfulness to 
retrieved evidence without gold answers \citep{es-etal-2024-ragas}, while 
RAGChecker measures claim-level coverage against a reference answer 
\citep{ru2024ragchecker}. Factuality-oriented alignment reduces hallucinations
\citep{tian2024finetuning,lin2024flame,chen2025train},
but optimizing factual precision alone can reduce
answer detail or relevance \citep{chen2025learning}. LLM-scored 
rubrics extend reward design to less easily verifiable properties 
\citep{gunjal2026rubrics}. Our reward measures coverage only over 
operator claims verified from sources available to the agent, and 
precision against evidence retrieved in its own trajectory.

\paragraph{Learning from better-informed demonstrators.}
Expert actions can be unreliable training targets when they depend
on information the learner cannot observe
\citep{weihs2021bridging}.
LLM agents can learn from privileged teacher feedback
\citep{choudhury2025better} or from teachers conditioned on
demonstration actions \citep{pmlr-v306-penaloza26a}.
Even within a single operator reply, only some claims may be
verifiable from the agent's sources.

\paragraph{Knowledge from support conversations.}
Support conversations yield question--answer pairs
\citep{laskar-etal-2025-ai} and service workflows
\citep{choubey2025turning}. Other work reformats policies
for LLMs \citep{su-etal-2025-llm} or tests whether agents retrieve
knowledge-base entries derived from user corrections
\citep{hazoom-etal-2026-iterate}. \citet{spangler2001knowledge}
identify gaps by matching ticket clusters to solution documents.
\citet{zhao-etal-2025-agent} ask operators to flag missing knowledge
during live support. Walker checks individual operator claims against
available rules and customer records to identify gaps in documentation,
tool access, or response rules.

\paragraph{Support agents and agentic retrieval.}
Modular support assistants separate intent detection, retrieval,
and response generation \citep{obadinma-etal-2022-bringing}. Our
agent uses retrieval-augmented generation
\citep{NEURIPS2020_6b493230} in a ReAct-style tool loop
\citep{yao2023react}.
Customer-support benchmarks evaluate tool use,
policy compliance, and knowledge retrieval
\citep{yao2025taubench,barres2025tau,shi2026tau}.
Production frameworks connect context engineering,
evaluation, and online testing \citep{gupta2026building},
while production-driven post-training uses observed
failure modes to inform task-specific reward design
\citep{tsymboi-etal-2026-production}.
GEPA studies reflective prompt optimization 
without updating model weights \citep{agrawal2026gepa}; our separate 
prompt-evolution baseline uses ShinkaEvolve~\citep{lange2025shinka} within SkyDiscover~\citep{liu2026skydiscover}.

\section{Framework}
\label{sec:system}

The pipeline combines four components (Figure~\ref{fig:pipeline}).
An offline verification agent, Walker, checks operator replies
against available evidence and uses unresolved claims to identify
gaps in the knowledge base (KB) and tools. A tool-using agent consults the revised KB
and customer records, while RL training rewards
grounded answers and coverage of verified operator claims.
Offline and online evaluation with expert review
assesses the resulting policy.
 
\paragraph{Problem Setting.}
At the latest customer turn, the agent receives the conversation $x$ and the request date. It 
may retrieve customer records and service rules before returning an outcome $y \in 
\{\textsc{Answer}, \textsc{Clarify}, \textsc{Handoff}, \textsc{ExitIntent}\}$. Answers must follow 
disclosure rules and be grounded in information available to the agent. Historical operator 
replies provide supervision but may contain information the agent cannot access. We define 
automation as the share of conversations closed without a human operator and monitor repeat 
contact separately, since closure does not necessarily mean resolution. Our deployment objective 
is to increase automation while keeping the share of human-reviewed production segments with a 
critical consultation error below 5\%.

\paragraph{Knowledge base and verification.}
We use recorded operator responses to identify the
knowledge and tools needed to resolve customer requests.
Walker, a GPT-5.2 \citep{gpt52} agent in the Codex harness \citep{codex},
decomposes replies into atomic claims and checks them
against retrieved KB passages and recorded tool results.
It records supporting sources, labels claims by evidence
support, and groups unresolved claims into knowledge,
customer-data, and servicing-rule gaps. Editors and developers review the findings and use them
to revise documentation, extend tool access, and clarify
response rules. The resulting KB contains 47K tokens across 12 files and was checked against 8680
factual claims extracted from 2760 historical operator
replies (App.~\ref{app:kb}).

\paragraph{Agent.}
The agent uses a single tool-and-observation loop, related to the ReAct pattern 
\citep{yao2023react}, rather than following manually authored dialogue flows. Its persistent context contains service instructions,
domain rules, and an index of KB documents with
brief descriptions of their contents. Seven read-only tools provide access to customer
application records and offers, look up applications
missing from the customer's list, and search and read
KB documents and FAQ entries using lexical queries.
After each tool result, the model chooses another
call or returns a typed outcome. For \texttt{Answer}, the agent drafts a reply, checks it against
the retrieved evidence and response rules, and then produces the final
answer. We use this format both to collect teacher trajectories for
SFT and to generate rollouts throughout GRPO training. The harness rejects invalid tool arguments and
a third identical tool call, returning corrective
feedback to the model (App.~\ref{app:agent}).
 
\paragraph{Training.}
We train the Qwen3.6-27B \citep{qwen3.6-27b} model using GRPO \citep{grpocite} with LoRA \citep{lora} on 4270
historical support conversations, using operator replies
as references and customer-data snapshots for offline
replay.
For responses containing text, the reward combines factual precision $p$ against
the agent's retrieved sources and intent-adjusted coverage $c_{\mathrm{eff}}$ of verified
operator claims to encourage informative answers
supported by the available evidence:
\[
R = \min\bigl(1,\,
p\underbrace{(0.3+0.7c_{\mathrm{eff}})}_{\text{coverage factor}}
+r_{\mathrm{ret}}\bigr)
\underbrace{w_{\mathrm{rule}}\,w_{\mathrm{traj}}}_{\text{penalty factors}}.
\]
Here, $r_{\mathrm{ret}}$ is a precision-weighted retrieval component,
$w_{\mathrm{rule}}$ penalizes response-rule violations
and intent misses, and $w_{\mathrm{traj}}$ penalizes
excessively long or malformed trajectories. Heuristic checks also limit rewards for inappropriate
non-text outcomes, while responses without factual
claims receive reduced precision credit
(App.~\ref{app:grpo}).  We compare GRPO with SFT on rewritten operator replies
and filtered teacher trajectories (App.~\ref{app:sft}).

\paragraph{Evaluation.}
We combine offline benchmarking with controlled online
tests and expert review. For offline evaluation and model selection, we use 500 randomly sampled production conversations
that the existing bot did not automate and a human operator
handled, excluded from training (App.~\ref{app:bench}). Our primary safety measure is the hallucination rate: the share of text
answers that a GPT-5.5 judge flags for an unsupported assertion, a
distortion or contradiction of the agent's evidence, or a violation of
a response rule. Because a low rate can be reached by answering less,
we also report intent match, the share of outcomes that address the
customer's need, and the rates of each outcome type. Both judges were calibrated against human annotations
(App.~\ref{app:judge_cal}). For online, candidates are compared in customer-level tests
against the existing production bot by automation
and operator time. Professional support-quality assessors review production conversations for critical consultation errors (App.~\ref{app:online}).

\section{Offline Experiments}
\label{sec:offline}

\paragraph{Main results.}
Direct reward optimization produced a more grounded
and more responsive agent than teacher imitation
(Table~\ref{tab:main-offline}).
The final GRPO agent had the lowest hallucination rate
among text answers at 8.6\% and the highest
text-answer rate at 88.9\%. Unlike SFT filtering, which reduced hallucinations
at the cost of fewer text answers, GRPO improved
grounding while answering more often.
Prompt evolution alone left most text answers flagged
for hallucinations, highlighting the value of
training in this setting (App.~\ref{app:evolution}). These gains nevertheless came with lower intent match
than SFT and the teacher, showing that stronger grounding
does not automatically yield more useful answers.
The following questions examine how supervision,
retrieval, and reward design shape this tradeoff.

\begin{table}[t]
\centering
\small
\setlength{\tabcolsep}{4pt}
\begin{tabular}{@{}lccc@{}}
\toprule
Model & Halluc.\,$\downarrow$ & Intent\,$\uparrow$ & Answer \\
\midrule

Qwen3.6-27B               & 68.6 & 89.0 & 72.3 \\
\quad + evolved prompt    & 58.6 & 84.8 & 70.9 \\
GPT-5.5                   & 17.5 & 86.0 & 71.7 \\
\quad + self-critique     & 15.8 & 82.9 & 70.9 \\
\midrule[0.3pt]
SFT                    & 13.6 & 80.4 & 60.2 \\
SFT$\to$GRPO           &  9.4 & 74.2 & 77.0 \\
GRPO (style)           & 23.5 & 72.4 & 78.2 \\
\textbf{GRPO (final)}  &  8.6 & 71.5 & 88.9 \\

\bottomrule
\end{tabular}
\caption{Offline results. Hallucination rate (Halluc.),
intent match (Intent), and text-answer rate (Answer)
are percentages. SFT used filtered GPT-5.5 trajectories with self-critique.
The GRPO (style) reward included
operator-template matching. Bold marks the deployed checkpoint.}
\label{tab:main-offline}
\end{table}

\paragraph{Can operator replies serve as references?}
To test whether recorded operator replies could serve as training
references, we used Walker to check their factual claims and
questions against the KB and recorded tool outputs. Walker verified
74.8\% of 8680 factual claims, but only 54.8\% of the 2760 replies
had no unresolved claims or questions. About half of the unresolved
claims concerned missing or incomplete customer records (App.~\ref{verifwalker}).

To make these replies useful as SFT targets, we asked a model
to reformulate each reply in the operator's style using only
claims supported by the annotated evidence.
The resulting agent matched the customer's intent in 55.2\% of cases,
compared with 80.4\% for SFT on filtered teacher trajectories (App.~\ref{app:sft}).
We therefore chose teacher trajectories for SFT. In GRPO, operator
replies serve as partial references: only claims supported by the
agent's KB or tool records count toward coverage.

\paragraph{When does retrieval reduce hallucinations?}
To discourage early GRPO policies from answering without consulting
the required KB sections, we varied the retrieval component weight. At 0.2, the hallucination rate among text answers
fell from 32.5\% to 26.7\% relative to no bonus, although the agent
also answered fewer requests in text. Doubling the weight to 0.4
raised the rate to 30.7\%, so we retained 0.2 (App.~\ref{app:grpo-judge}).

After adding an incentive to search, we tested whether
better search results would further reduce hallucinations. On 207
KB-dependent conversations, replacing the final GRPO agent's lexical
search results with gold KB data changed its hallucination rate
from 8.3\% to 8.2\%, with no gain in intent match. Giving GPT-5.5
the entire 47K-token KB likewise had little effect, lowering its hallucination rate from 17.5\% to 17.1\%.
Across GPT-5.5, Qwen3.6-27B, and GRPO, KB queries were
typically short keywords, and 69--78\% concerned purchase categories
and merchants, eligibility exclusions, amounts and deadlines, bonuses,
or term definitions. Given the small gains from supplying evidence
directly, we retained lexical search for its simplicity.

\paragraph{Does structured self-critique help?}

Structured self-critique provides an explicit record
of answer revision for SFT without access to the
teacher's internal reasoning. We asked GPT-5.5 to draft a reply, check
it for unsupported claims and violations of response rules,
and then produce the final answer (App.~\ref{app:agent}).
We tested whether this revision step improved the teacher's answers
by running GPT-5.5 with and without it (Table~\ref{tab:main-offline}).
Across 3 runs, self-critique reduced the hallucination rate
from 17.5\% to 15.8\% (paired 95\% CI for the reduction,
0.6--3.3 p.p.). The text-answer rate stayed near  71\% in both settings, while intent match fell from 86.0\% to 82.9\%.

\paragraph{How should the reward balance precision and coverage?}
To encourage informative answers without rewarding unsupported claims,
we compared precision-only, coverage-only, and $F_1$ rewards
(Table~\ref{tab:precision-coverage}).
Precision $p$ measures grounding against evidence retrieved in the
agent's own trajectory. Coverage measures reproduction of operator
claims, using either all claims ($c_{\mathrm{all}}$) or only those
verified from the agent's available sources ($c_{\mathrm{supp}}$).

\begin{table}[t]
\centering
\small
\setlength{\tabcolsep}{4pt}
\begin{tabular}{@{}lrrr@{}}
\toprule
Reward & Halluc.\,$\downarrow$ & Intent\,$\uparrow$ & Answer \\
\midrule
Precision $p$                  & 24.8 & 61.8 & 77.6 \\
Coverage $c_{\mathrm{supp}}$    & 76.7 & 86.8 & 81.2 \\
$F_1(p,c_{\mathrm{all}})$       & 39.1 & 72.6 & 81.2 \\
$F_1(p,c_{\mathrm{supp}})$      & 32.5 & 68.3 & 81.2 \\
\bottomrule
\end{tabular}
\caption{Precision and coverage reward designs.
Hallucination rate (Halluc.), intent match (Intent),
and text-answer rate (Answer) are percentages.
Coverage targets include either all operator claims
or only verified claims.}
\label{tab:precision-coverage}
\end{table}

Coverage alone left unsupported additions unpenalized and produced
high hallucination rates, even with verified targets.
Precision alone yielded fewer hallucinations but lower intent match
and fewer text answers than $F_1$ with verified coverage.
Within $F_1$, restricting coverage targets to verified
operator claims reduced hallucinations at an
unchanged text-answer rate, although intent match also decreased.
These comparisons expose a tradeoff: rewarding more reference content
can improve intent match while weakening grounding.

The final reward uses a precision-dominant core
$p(0.3+0.7c_{\mathrm{eff}})$, where $c_{\mathrm{eff}}$ accounts for
intent fulfillment (App.~\ref{app:grpo-final}).
The $0.3p$ term preserves credit for grounded answers even when
they cover none of the verified reference claims, whereas the
$F_1$ core gives zero credit at zero coverage. With the retrieval component and rule penalty held fixed, switching cores
lowered hallucinations from 17.2\% to 12.3\%. Intent
match rose by 1.4~p.p. and the text-answer rate fell by 1.6~p.p.
(Table~\ref{tab:reward-constraints}).

\paragraph{How do expert feedback and serving constraints shape the reward?}
To turn recurring consultation errors into training
signals, we first added a \emph{rulebook gate} to
the $F_1$ reward with the retrieval bonus.
The gate multiplied the reward by $0.4+0.6g$, where $g$ is the
judge's score for compliance with the rules in the agent's system prompt. The offline hallucination judge already counted violations
such as disclosing the inviter's bonus to the invited customer.
Giving these violations an explicit training penalty lowered the
overall hallucination rate from 26.7\% to 17.2\%
(Table~\ref{tab:reward-constraints}).

Professional support-quality assessors reviewed
674 agent answers, from which we derived 14 binary
checks (App.~\ref{app:rubrics}). Among the errors assessors identified were confusing the inviter
with the invited friend, omitting when a bonus would be paid, and
ignoring a customer's frustration. We retained ten checks that flagged our fine-tuned
agent's responses in more than 5\% of dialogues
and fired significantly more often than on
operator replies. Each check received a weight of 0.1--0.5, and we combined the resulting flags with a separate penalty
for failing to address the customer's intent.
With the precision-dominant core held fixed, replacing the rulebook
gate with this penalty raised intent match by 10.0~p.p.
The text-answer rate rose from 78.2\% to 89.3\%, while the
hallucination rate among those answers fell from 12.3\% to 8.1\%.

\begin{table}[t]
\centering
\small
\setlength{\tabcolsep}{3pt}
\begin{tabular}{@{}llrrr@{}}
\toprule
Core & Penalty & Halluc.$\downarrow$ & Intent$\uparrow$ & Answer \\
\midrule
$F_1$        & None                 & 26.7 & 60.2 & 70.0 \\
             & Rulebook gate        & 17.2 & 59.1 & 79.8 \\
\midrule[0.3pt]
$p$-dominant & Rulebook gate        & 12.3 & 60.5 & 78.2 \\
             & Weighted rubrics     &  8.1 & 70.5 & 89.3 \\
             & \shortstack[l]{\textbf{Rubrics}\\\textbf{+ trajectory}} 
             & 8.6 & 71.5 & 88.9 \\
\bottomrule
\end{tabular}
\caption{Offline results for GRPO reward variants with verified
coverage and a shared retrieval component. Hallucination rate (Halluc.), intent match (Intent),
and text-answer rate (Answer) are percentages.
The bold label marks the final reward.}
\label{tab:reward-constraints}
\end{table}

Serving constraints motivated additional penalties for
malformed outputs and excessive generation in the final reward. We set the soft length
threshold at 5K total generated tokens per trajectory, the 99th
percentile for GPT-5.5 teacher trajectories. Adding these penalties
changed each offline metric by at most 1 p.p.
(Table~\ref{tab:reward-constraints}). At inference, we cap reasoning
at 5K tokens based on a separate comparison of answer quality and
generation length (Table~\ref{tab:reasoning}).

\section{Online Experiments}
\label{sec:online}
 
\begin{table}[t]
\centering
\small
\setlength{\tabcolsep}{2.75pt}
\begin{tabular}{@{}lcccc@{}}
\toprule
Arm & $N$ & Automation & $\Delta$ Op. time/user & Crit. \\
\midrule
\multicolumn{5}{@{}l}{\emph{Candidate-selection trial, 9 days}} \\
Procedural bot & 3789 & 63.3\% & 0.0\%   & --- \\
GPT-5.5        & 3919 & 85.2\% & -29.2\% & --- \\
SFT            & 3821 & 82.2\% & -26.9\% & 3.1\% \\
GRPO (style)   & 3778 & 78.6\% & -11.0\% & 2.8\% \\
\textbf{GRPO} (final)   & 3811 & 84.4\% & -28.6\% & 0.9\% \\
\midrule
\multicolumn{5}{@{}l}{\emph{Product A/B, 8 days}} \\
Procedural bot & 17552 & 63.3\% & 0.0\%   & --- \\
\textbf{GRPO (final)}   & 17745 & 80.8\% & -30.4\% & 0.7\% \\
\bottomrule
\end{tabular}
\caption{Online tests reporting automation, percentage change
in average operator time per user relative to the procedural bot ($\Delta$ Op. time/user),
and critical-error rates in expert-reviewed samples (Crit.).
$N$ counts customers in the candidate-selection trial and conversations
in the product A/B.}
\label{tab:main-online}
\end{table}

\paragraph{Setup and results.}
We sought an in-house agent to control inference costs and avoid
relying on an external API in production. In a nine-day randomized
candidate-selection trial involving 19118 customers, we compared
SFT and two GRPO variants with GPT-5.5 and the production bot
(Table~\ref{tab:main-online}). Final GRPO achieved the highest
automation among the in-house candidates at 84.4\% and required
less operator time per user than SFT. Experts found critical errors in 0.9\% of reviewed conversations for final GRPO and
3.1\% for SFT. The 95\% confidence intervals did not overlap. We therefore selected
final GRPO for the product A/B test.

In the eight-day product A/B test against the production bot,
automation rose from 63.3\% to 80.8\% (+17.5 p.p., 95\% CI 16.4--18.8) and operator time per user
fell by 30.4\% (95\% CI 27.1--34.3). Repeat contact within 48 hours also fell, from
34.1\% to 28.9\%. Experts found at least one consultation error
in 9.3\% of reviewed interactions with GRPO and 25.9\% of those with the procedural bot
($p<0.001$). A separate review of 722 GRPO segments found a
critical-error rate of 0.7\% (95\% CI 0.3--1.6\%), below the 5\%
acceptance threshold (App.~\ref{app:online}).

\paragraph{Do offline helpfulness proxies predict automation?}
Not necessarily. We tested whether matching operators in what to
tell customers and how to present it predicted higher automation. Operator-answer templates provided a common reference for cases
within the same scenario.  We selected templates whose required
content could be grounded in the agent's sources. Case-specific values were replaced by
placeholders, leaving factual correctness to the precision reward.
We calibrated the template-conformity judge against human annotations.
On a separate held-out sample, GRPO (style) won
74.2\% of non-tied template-guided pairwise comparisons with GRPO (final). The mean pointwise
\texttt{style\_success} score was 0.494 for the style variant and 0.389 for the final variant
(App.~\ref{app:grpo-proxies}).
In the nine-day candidate-selection trial (Table~\ref{tab:main-online}), automation was 78.6\% for GRPO (style) and 84.4\% for GRPO (final).
Calibrating the pairwise judge against human preferences therefore
does not replace validating the proxy against online automation.

\paragraph{Do SFT and GRPO agents fail in different ways?}

To compare grounding failures in production, we examined \emph{unsupported factual claims} flagged by assessors during the same review that produced the critical-error rates
in Table~\ref{tab:main-online}. On the same 800 conversations per model,
assessors checked whether facts stated in an answer, such as bonus
amounts, deadlines, or eligibility conditions, were supported by the
KB passages and tool outputs available to the agent. They flagged
unsupported claims regardless of whether the error was critical.
Such claims appeared in 5.0\% of SFT answers and 0.6\% of final
GRPO answers (App.~\ref{app:candcomparison}).

This difference is consistent with how the two agents were trained.
SFT learns to reproduce the teacher's tool calls and answers.
In GRPO, the factual precision reward directly penalizes claims
that lack support in the agent's own trajectory.

\section{Conclusion}
We trained and deployed a referral-support agent using historical
operator replies. Some replies relied on information the agent
could not access. We checked individual operator claims against
the agent's KB and customer records. Unresolved claims guided KB and
tool revisions, while verified claims defined coverage targets for
GRPO. The agent earned factual grounding credit for claims supported
by evidence it retrieved in the current conversation. We selected
the deployment candidate on live traffic, where both offline measures
of template conformity had favored the candidate that automated fewer
conversations. In the product A/B test, the selected system increased
automation by 17.5 p.p. over the procedural bot and
reduced operator time per customer. Expert review found its
critical-error rate below the deployment threshold. 
The same training
approach can be applied in support workflows where historical operator
replies can be paired with the rules and records available to the
agent at the time of each request.

\section*{Limitations}
\paragraph{Scope and reproducibility.}
We evaluate the approach in one language and a single
customer-support domain, referral programs, within
one organization. Training experiments use only
Qwen3.6-27B. The agent retrieves information but does
not perform servicing actions, so the results do not
establish reliability for agents with broader operational
authority. Proprietary customer records and reliance
on external models limit independent reproduction.

\paragraph{Evaluation coverage.}
The offline benchmark contains requests that the previous
bot handed off to operators and evaluates one response
without continuing the dialogue. It therefore does not
represent the full production traffic or directly measure
end-to-end resolution. Hallucination assessment covers
text answers but not factual assumptions embedded in
clarification questions. Despite human calibration and consistent SFT--GRPO
ranking across four judges, absolute hallucination
rates remain sensitive to the judge and rubric. Grounding checks establish support in available sources,
whose correctness and completeness are not guaranteed. Repeated use of the offline benchmark for reward
and checkpoint selection may make its reported
performance optimistic. Independent online experiments
validate the selected system's operational performance.

\paragraph{Deployment evidence.}
The product A/B test evaluates the complete system,
including its KB, tools, harness, and trained
policy, without isolating their individual contributions.
The controlled experiment lasted eight days, followed
by approximately five weeks of monitoring. These results
do not establish long-term robustness to changes in
customer requests, product rules, or back-end systems.

\section*{Ethical Considerations}
\paragraph{Intended use and risks.}
The agent is intended for customer support within a defined
service scope and runs on in-house infrastructure. Its tools
retrieve information but do not perform servicing actions.
Requests requiring additional authority are handed off to
human operators. Grounding checks and response rules reduce
but do not eliminate the risk of incorrect or misleading
answers.

\paragraph{Human oversight and operator impact.}
Deployment decisions combine offline evaluation, controlled
online experiments, and expert assessment of critical
consultation errors under a strict review protocol. Trained assessors
continue to monitor consultation quality after deployment. We evaluate support operator time
alongside consultation quality, but do not assess the
longer-term effects of automation on employment,
workload distribution, or working conditions.

\paragraph{Data privacy.}
All personal data were masked before any calls to
external models, including names, contact details, customer identifiers and other personal
information. Masking used synthetically generated substitutes
to preserve data consistency without exposing
sensitive information.

\bibliography{custom}
\clearpage

\appendix

\section{Knowledge Base Construction}
\label{app:kb}

\subsection{Initial Knowledge Base}

The initial knowledge base was assembled in consultation with the product team, using publicly available company sources. The collection was subsequently extended with customer-facing help articles and text extracted from operator procedures. At the first verification stage, it already contained ten product-specific rule documents in Markdown and a source map.

The initial content comprised approximately 34K tokens across ten files. The revised collection contained approximately 47K tokens across twelve files: the ten rule documents, a structured FAQ collection, and an editorial reference document. Counts use the \texttt{o200k\_base} tokenizer and exclude conversation data and tool records.

\subsection{Knowledge Base Schema}
The knowledge base contains three types of content:
\begin{itemize}
    \item \textbf{Product-specific rule documents.} Program terms, eligibility criteria, and participation requirements.
    
    \textit{Example:} Rules for the mobile-service referral offer.

    \item \textbf{Structured FAQ entries.} Questions and answers with product metadata and source URLs.
    
    \textit{Example:} ``Can I receive a reward if I invited a friend without a referral link?''

    \item \textbf{Editorial reference cards.} Terminology, interface navigation, and common service scenarios.
    
    \textit{Example:} ``Where is the referral section in the app and online account?''
\end{itemize}
A source map identifies each document's scope and specifies source-selection and precedence rules.

\subsection{Verification with Walker}
\label{verifwalker}

Historical operator replies do not record the provenance of individual assertions. A reply may combine program rules, customer-specific records, and procedural knowledge, some of which is unavailable to the serving agent. Walker, an offline verification agent based on GPT-5.2 and running in the Codex harness, checks these assertions against the knowledge base and available tool records.

Walker extracts atomic factual claims and questions from each operator reply. It verifies claims against relevant documents and recorded tool results, checks whether questions are justified by the available context, and records supporting source locations. Items receive the labels \texttt{supported}, \texttt{partially\_supported}, \texttt{not\_found}, or \texttt{not\_checked}. Items labeled \texttt{not\_checked} are excluded from evaluation.

We reported three diagnostic measures: the share of operator factual claims supported by the available evidence ($C_{\mathrm{claim}}$), the share of operator questions justified by the available context ($C_{\mathrm{query}}$), and the share of replies with no unresolved items ($C_{\mathrm{reply}}$).

Let $\mathcal{D}$ denote successfully processed replies, $\mathcal{A}_d$ their checked items, and $\mathcal{C}_d$ and $\mathcal{Q}_d$ their checked factual claims and questions. For status $s(x)\in\{S,P,U\}$, denoting supported, partially supported, and not found, respectively:

\[
C_{\mathrm{claim}} = \frac{\sum_{d\in\mathcal{D}}\sum_{x\in\mathcal{C}_d}\mathbf{1}[s(x)=S]}{\sum_{d\in\mathcal{D}}|\mathcal{C}_d|}.
\]

\[
C_{\mathrm{query}} = \frac{\sum_{d\in\mathcal{D}}\sum_{x\in\mathcal{Q}_d}\mathbf{1}[s(x)=S]}{\sum_{d\in\mathcal{D}}|\mathcal{Q}_d|}.
\]

\[
C_{\mathrm{reply}}
= \frac{1}{|\mathcal D|}
  \sum_{d\in\mathcal D}
  \mathbf{1}\!\left[\forall x\in\mathcal A_d,\; s(x)=S\right].
\]

Combined item support, $C_{\mathrm{item}}$, uses $\mathcal{A}_d$ in place of $\mathcal{C}_d$ in the claim-support formula.
Query justification diagnoses clarification behavior, including unnecessary requests for identifiers and questions containing unsupported customer-specific assumptions. Flagged questions enter the revision report and affect reply-level support.

The verification protocol evolved during development. Early prompts excluded clarification questions; later prompts evaluated their justification, with complete claim/question annotations available in the revised-KB run. Table~\ref{tab:kb-runs} leaves unavailable separate scores unreported.

\begin{table*}[t]
\centering\small
\setlength{\tabcolsep}{3.5pt}
\begin{tabular}{@{}lrrrrrrr@{}}
\toprule
Run & \shortstack{KB tokens\\(thousands)} & \shortstack{Replies\\$|\mathcal{D}|$} & \shortstack{Checked items\\$\sum_{d\in\mathcal{D}}|\mathcal{A}_d|$} & \shortstack{Item support\\$C_{\mathrm{item}}$} & \shortstack{Claim support\\$C_{\mathrm{claim}}$} & \shortstack{Query justification\\$C_{\mathrm{query}}$} & \shortstack{Reply support\\$C_{\mathrm{reply}}$} \\
\midrule
Pilot & 34.1 & 200 & 780 & 60.0\% & --- & --- & 46.0\% \\
Expanded sample & 43.6 & 1230 & 5040 & 64.9\% & --- & --- & 43.8\% \\
Draft KB & 48.3 & 1090 & 5400 & 62.8\% & --- & --- & 30.3\% \\
Revised KB & 46.9 & 2760 & 9210 & 75.9\% & 74.8\% & 93.2\% & 54.8\% \\
\bottomrule
\end{tabular}
\caption{Stages of knowledge-base development guided by Walker
runs, from the initial pilot to the revised collection.}
\label{tab:kb-runs}
\end{table*}

The revised-KB run evaluated 8680 factual claims and 530 questions. Table~\ref{tab:kb-gap-types} summarizes the categories assigned to factual claims without full support. 
Tool-data issues concern missing, incomplete, or ambiguous customer-specific evidence, such as the relevant application, participant, reward, or deadline. 
Knowledge-document gaps concern general rules or instructions not found in the available documents. 
Servicing-rule issues concern procedural assertions, such as whether a decision can be reconsidered or when support will respond. These categories describe Walker's diagnosis of unresolved items.

\begin{table}[t]
\centering\small
\begin{tabular}{@{}lr@{}}
\toprule
Reported issue category & Factual claims \\
\midrule
Tool-data issues & $\sim$1130 \\
Knowledge-document gaps & $\sim$580 \\
Servicing-rule issues & $\sim$480 \\
\midrule
Total without full support & $\sim$2190 \\
\bottomrule
\end{tabular}
\caption{Reported issues in the revised-KB run.}
\label{tab:kb-gap-types}
\end{table}

Walker was calibrated on batches of up to 30 examples, each manually reviewed by the developer. Its instructions were adjusted to prioritize recall for unsupported and partially supported claims (\texttt{not\_found} and \texttt{partially\_supported}), accepting false alarms to reduce missed evidence gaps. Evaluation on 100 examples shows a precision of 0.66 and a recall of 0.90 for detecting these gaps.
This calibration favors finding potential gaps, but a flagged claim does not necessarily require a KB update. The evidence may already be present but missed by Walker, or the operator's statement may be incorrect. Manual review suggested that operator errors affected at most approximately 5\% of inspected examples, although their prevalence in the full corpus was not measured. Flagged claims therefore require review before they are used to revise the KB, and coverage alone cannot establish whether a revision improves the serving agent.

\subsection{Revision Loop}

\begin{figure*}[t]
\centering
\includegraphics[width=\textwidth]{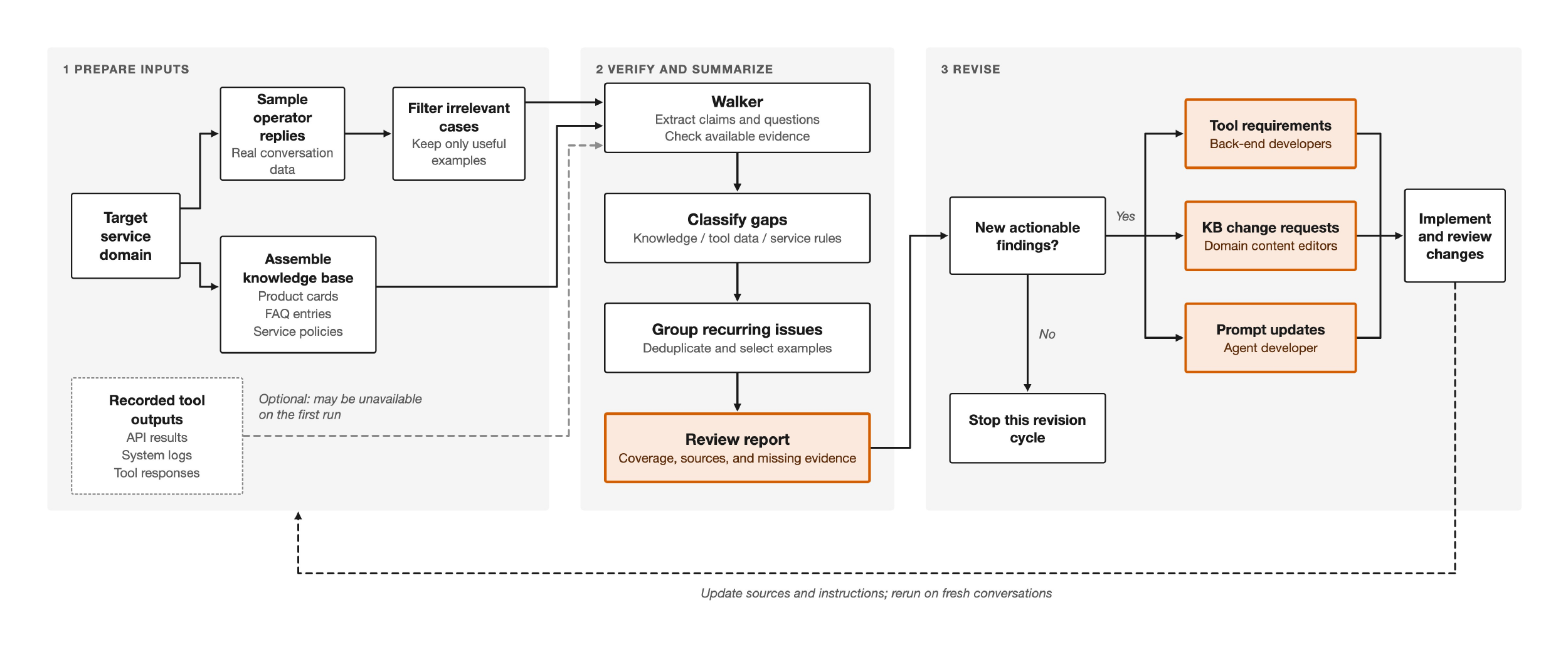}
\caption{Walker verification and revision loop. Structured verification results are classified and grouped into recurring issues. Report review produces requests for tool development, content revision, or prompt updates. Verification is repeated on fresh conversations after changes are implemented and reviewed. Tool records may be unavailable on the first run.}
\label{fig:walker-revision-loop}
\end{figure*}

Figure~\ref{fig:walker-revision-loop} summarizes the verification and revision workflow. Operator replies are sampled within the selected service domain, and irrelevant cases are filtered out. Walker checks the remaining replies against the KB and available tool records. Tool records may be unavailable on the first run; the resulting gaps help identify required tool capabilities.

Figure~\ref{fig:kb-revision-example} shows an excerpt from an actual Walker verification record. Walker marked the claim about a 96-hour status-update period as \texttt{not\_found}, leaving \texttt{supporting\_facts} empty. The addition provides the missing rule using wording from the current KB. Subsequent classification assigns \texttt{problem\_type} and \texttt{problem\_type\_reason} to unresolved items before they are grouped for review.

\begin{figure*}[t]
\centering
\begin{minipage}{0.95\textwidth}
\small
\textbf{Walker output (excerpt, translated).}
\begin{Verbatim}[fontsize=\small, frame=single]
{
  "row_index": 3310,
  "claims_to_facts": [
    {
      "operator_answer_locator":
        "Information on requirement completion is updated within 96 hours.",
      "claim":
        "The promotion's completion status is updated within 96 hours.",
      "is_query": false,
      "status": "not_found",
      "note":
        "The consulted sources and available tool outputs do not
         confirm the 96-hour update period.",
      "supporting_facts": []
    }
  ],
  "all_facts_found": false
}
\end{Verbatim}

\textbf{Editorial finding.}
The missing evidence concerns the delay between completing the promotion requirements and the corresponding status update, rather than the reward payment deadline.

\medskip
\textbf{Editor addition to the KB.}
In the card on completed requirements with no reward received:
\begin{quote}
After all promotion requirements have been completed, the status is updated within 96 hours.
\end{quote}
\end{minipage}
\caption{A reported evidence gap concerning a status-update delay. The Walker excerpt comes from an actual verification result.}
\label{fig:kb-revision-example}
\end{figure*}

Items labeled \texttt{partially\_supported} or \texttt{not\_found} are classified according to the evidence or guidance needed to resolve them:
\begin{itemize}
    \item \textbf{Knowledge-document gaps:} missing general product rules, eligibility conditions, or reusable instructions.
    \item \textbf{Tool-data issues:} missing customer-specific facts, incomplete tool records, or insufficient links between an application, participant, and reward.
    \item \textbf{Servicing-rule issues:} missing guidance on service decisions or communication, such as whether a decision can be reconsidered or which information may be disclosed.
\end{itemize}
Classification uses the extracted item, the verifier's explanation, and any supporting evidence. For partially supported items, the report identifies the missing part rather than requesting information already supported by the sources. Questions are reviewed separately, and evidence available only from unverified sources is flagged for confirmation.

During report generation, the agent groups items that describe the same missing information or required correction. Deduplication consolidates repeated cases into a shared issue description while retaining distinct subproblems and the underlying examples. Each group records the number of affected items and conversations. The full set of examples remains accessible in expandable report sections.

For each KB group, the report presents five representative examples. A domain content editor reviews the group and these examples, consulting additional cases when needed to clarify its scope. The editor then searches for the missing information in the KB and other available company documents.

If the information is already present in the local KB, the editor supplies the relevant document or passage, identifying evidence missed by Walker. If an existing document is absent from the local collection, it is supplied for inclusion. If no documented answer exists, the editor consults representatives of the relevant domain teams, clarifies the rule or process, and records the required information in an appropriate knowledge source.

Report review also produces tool requirements for back-end developers and prompt updates for the agent developer. The responsible specialists implement and review the changes, after which Walker is rerun on fresh conversations using the updated sources and instructions.

The number of runs depends on the initial availability of sources and tools. The process may require only two runs: an initial run without tool data to identify missing capabilities, followed by a run with tool outputs once those capabilities become available. Further runs are performed as needed. Iteration stops when additional runs yield no new insights, rather than when coverage reaches 100\%.

Table~\ref{tab:kb-effort} summarizes the reported effort and duration of the main activities. Pipeline preparation depends on the difficulty of obtaining conversation exports. Report analysis takes approximately one working day. Editorial revision was estimated at five days for this case, while tool development was a separate setup activity completed in approximately one week.

\begin{table}[t]
\centering\small
\begin{tabular}{@{}ll@{}}
\toprule
Activity & Reported time \\
\midrule
Pipeline preparation and launch & 1--5 hours \\
Report analysis & 8 hours \\
Editorial revision & $\sim$5 days (estimated) \\
Tool development & $\sim$1 week \\
\bottomrule
\end{tabular}
\caption{Approximate effort and duration by activity.}
\label{tab:kb-effort}
\end{table}

\section{Agent Architecture and Harness}
\label{app:agent}

\paragraph{From Dialogue Tree to Agent}
The original production bot was organized as a rule-based dialogue tree. Intent classifiers and regular expressions selected among options that narrowed down the customer's request; the selected branch then retrieved relevant customer records or returned a prewritten consultation. Routing, retrieval, and answering were configured in separate blocks, so extending the procedure required maintaining the transitions between them. An intermediate branch-constrained agent used an LLM for intent classification but retained predetermined paths and lacked the missing-application lookup in Table~\ref{tab:agent-tools}.

\begin{figure*}[t]
\centering
\includegraphics[width=\textwidth]{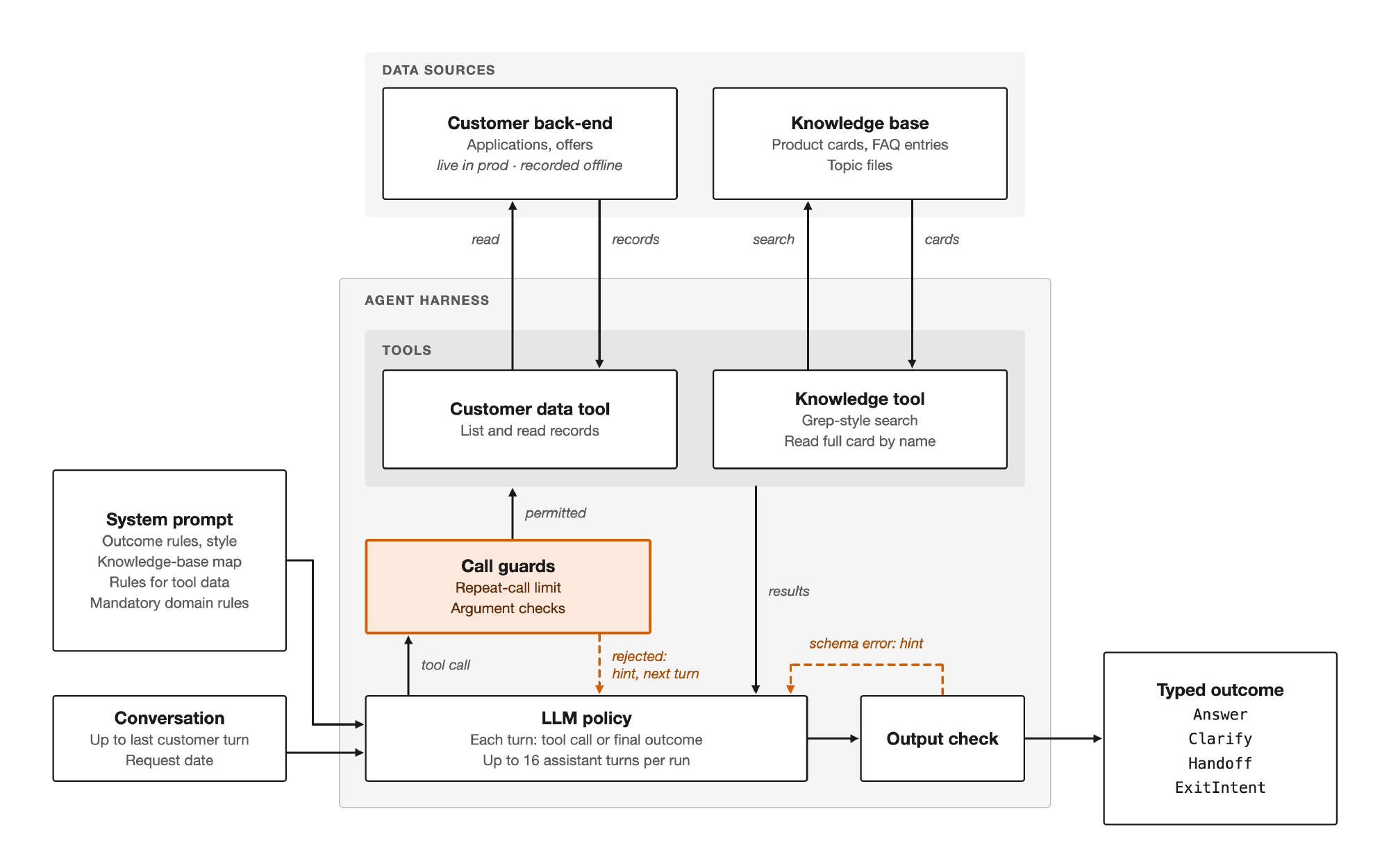}
\caption{Agent harness. The model receives the conversation up to the last customer turn, the request date, and a fixed system prompt. On each turn it either calls a tool or returns a typed outcome. Customer-data tools query live back-end records in production and use available recorded data offline; knowledge tools search static product cards and FAQ entries.}
\label{fig:agent-harness}
\end{figure*}

We subsequently adopted a flexible tool-using agent (Figure~\ref{fig:agent-harness}). The model receives the domain rules and available tools in one context, selects a tool, observes its result, and decides whether to call another tool or return a final outcome. Search and answer generation belong to this single decision loop. This design is simpler to extend because new rules and tool guidance do not require manually coordinating transitions among routing and consultation blocks.

\paragraph{Context and Tools}
The agent's persistent context contains basic service instructions, domain rules, a map of the knowledge base, and guidance on interpreting tool fields. It can retrieve customer-specific records and available offers through data tools. For general rules, it can search product cards and FAQ entries with a case-insensitive lexical search tool, then read the relevant part of a knowledge card. The model decides which source it needs for the current question and can alternate between data retrieval and knowledge-base search as the case develops. Table~\ref{tab:agent-tools} lists all tools with their inputs and outputs.

\paragraph{Knowledge Retrieval}
We use model-directed lexical search for the knowledge base. The model formulates a query, inspects matched passages with their source locations, and can read a card or refine the query. This keeps retrieval inspectable without maintaining a separate embedding model and vector index. In other agentic-search settings, grep-based retrieval has performed well relative to vector retrieval, although results depend on the harness and task~\citep{sen2026grepneedagentharnesses}. This motivates our interface; we do not infer a performance advantage over semantic retrieval on our own knowledge base.

\paragraph{Inference-Time Guards}
Tool calls pass through two inference-time guards, and the harness handles their rejections inside the decision loop rather than aborting the run. The argument guard rejects a call with invalid arguments, such as an unknown offer or card identifier or a value outside an enumeration. The repeat-call guard rejects a third call of the same tool with identical arguments, so that the model cannot loop on a query that returns nothing new. In both cases the tool returns a short textual hint instead of a result, and the model sees it on its next turn; for example, ``offer \texttt{X} is not among the available offers; call \texttt{list\_available\_offers} to see the existing identifiers'' or ``you have already called this tool with these arguments; a repeated call will return nothing new''. A final outcome that fails schema validation is returned to the model in the same way. Each tool and the final outcome allow one such correction in a row; a second consecutive rejection terminates the run as a model failure, and a run is capped at 16 model requests. A record that is absent for the current chat is not an error: the tool returns an empty result, and the absence is logged separately.

\begin{table*}[t]
\centering\small
\setlength{\tabcolsep}{5pt}
\renewcommand{\arraystretch}{1.15}
\begin{tabular}{@{}lp{11.8cm}@{}}
\toprule
Tool & Description \\
\midrule
\multicolumn{2}{@{}l}{\textit{Customer records}} \\[2pt]
\texttt{get\_applications\_if} & Takes no arguments. Returns the customer's referral applications in short form: application identifiers and both parties, the inviter and the invited friend. \\
\addlinespace[3pt]
\texttt{get\_application\_details} & Takes an optional \texttt{application\_id}. Returns detailed application cards: status, decline reason, progress steps, bonus amounts, payment status, and deadline; without an identifier, all cards of the customer. \\
\addlinespace[3pt]
\texttt{list\_available\_offers} & Takes no arguments. Returns the referral offers available to the customer: offer identifier, product, and target action. \\
\addlinespace[3pt]
\texttt{get\_offer\_details} & Takes an \texttt{offer\_id}. Returns the full terms of that offer: steps, bonus, deadline in days, referral link, and link to the legal terms. \\
\addlinespace[3pt]
\texttt{find\_missing\_application} & Takes the friend's name and phone, the product, and the customer's role. Searches the bank's back end for an application that did not link to the friend, typically because the friend applied on the website instead of through the referral link. Returns a reply template with the outcome and deadlines, which the agent quotes verbatim. \\
\midrule
\multicolumn{2}{@{}l}{\textit{Knowledge base}} \\[2pt]
\texttt{grep\_knowledge} & Takes a \texttt{query} and \texttt{max\_results} (default 7). Lexical retriever: a case-insensitive substring match of the query against single lines of product cards and against FAQ entries, with no embeddings or ranking. Returns up to four hits from product cards and up to three from the FAQ, each with the card name, line number, section heading, and a snippet of three lines (full entry for the FAQ). \\
\addlinespace[3pt]
\texttt{read\_knowledge\_card} & Takes a card \texttt{name}, a line \texttt{offset} (default 0), and a \texttt{limit} (default 2000 lines). Returns that line range of the card, usually the whole document. \\
\bottomrule
\end{tabular}
\caption{Tools available to the agent. Customer-record tools read back-end data for the current chat; knowledge-base tools search shared documents. All tools retrieve information and perform no servicing actions. In offline evaluation, every call is answered from recorded data.}
\label{tab:agent-tools}
\end{table*}

\paragraph{Final Outcomes}
The agent has four typed final outcomes: \texttt{Answer}, \texttt{Clarify}, \texttt{Handoff}, and \texttt{ExitIntent}. \texttt{Answer} is a customer-facing reply grounded in the retrieved records and knowledge-base content. The model writes a draft, critiques it against the style and grounding rules, and returns the revised text. \texttt{Clarify} asks the customer a question when the request cannot be answered yet, for example which of several applications is meant, and may offer suggested replies for the customer interface. The remaining two outcomes end the agent's part of the conversation and differ in where the dialogue goes next. \texttt{Handoff} transfers the dialogue to a human operator. It is returned when resolution requires authority or servicing actions beyond the agent's scope, when the customer explicitly asks for an operator, or when a dispute remains unresolved, for example when the customer rejects
the agent's explanation. A difficult question that the agent can still answer from its sources is not by itself a trigger. \texttt{ExitIntent} leaves the referral scenario and returns the dialogue to general routing when the customer has moved to an unrelated topic or the last message carries no usable request. The agent's tools retrieve information but do not perform servicing actions. Human operators can consult additional internal context and carry out actions unavailable to the agent.

\paragraph{Offline Replay}
In production, the agent receives the live conversation and its tools query current back-end data. For offline evaluation, the harness instead runs the agent on logged conversations and answers its tool calls with recorded data for the corresponding chat. The recorded data are a full export of the customer's back-end records at the time of the chat, so the tools return the same information the agent would have received in production, and the two settings do not differ in what the agent sees. It records the selected calls, their results, the intermediate trajectory, the typed final outcome, and execution failures. The recorded data make cases repeatable while allowing different agent versions to choose different sequences of tools.

\section{Offline Benchmark}
\label{app:bench}

\paragraph{Construction}
We constructed the benchmark from production conversations that the existing bot did not automate and that were handled by a human operator. We randomly sampled 500 conversations from this operator-handled stream; we did not stratify the sample. Each case includes the conversation, the operator's reply, and the recorded customer applications and offers available for offline replay. The benchmark is single-turn: the conversation is cut at the last customer message, the agent produces one outcome in place of the operator's reply, and the dialogue is not continued. The benchmark therefore tests both which information the agent chooses to retrieve and how it uses that information in its final response. Benchmark dialogues do not overlap with the training data: the SFT and GRPO dialogues were drawn from the same logs, and every benchmark dialogue was excluded from them. Training and evaluation share one harness (App.~\ref{app:agent}), so a candidate is scored under the same prompts, tools, and guards it was trained with.

\begin{table*}[t]
\centering\small
\setlength{\tabcolsep}{4pt}
\renewcommand{\arraystretch}{1.2}
\begin{tabular}{@{}p{2.0cm}p{3.3cm}p{6.6cm}cp{1.9cm}@{}}
\toprule
Metric & Formula & Definition & Better & Judge \\
\midrule
Hallucination rate &
$\displaystyle\frac{1}{|\mathcal{T}|}\textstyle\sum_{a\in\mathcal{T}}\mathbf{1}[h(a)=1]$ &
Share of text answers the judge flagged. $\mathcal{T}$ is the set of \texttt{Answer} outcomes. The judge sets $h(a)=1$ when answer $a$ contains an unsupported assertion, distorts or contradicts the evidence, or breaks a response requirement. &
$\downarrow$ & GPT-5.5, medium \\
\addlinespace[4pt]
Intent match &
$\displaystyle\frac{1}{|\mathcal{O}|}\textstyle\sum_{o\in\mathcal{O}}\mathbf{1}[g(o)=1]$ &
Share of outcomes that address the customer's need. $\mathcal{O}$ is the set of all final outcomes. The judge sets $g(o)=1$ when outcome $o$ has none of the intent errors
 listed in App.~\ref{app:intent-judge}: an answer or question off the customer's need, an ignored handoff request, or an unacknowledged negative sentiment. &
$\uparrow$ & GPT-5.5, medium \\
\addlinespace[4pt]
Outcome rates &
$\displaystyle\frac{|\mathcal{O}_t|}{|\mathcal{O}|}$ &
Share of each outcome type. $\mathcal{O}_t$ contains the outcomes of type $t$: \texttt{Answer}, \texttt{Clarify}, \texttt{Handoff}, or \texttt{ExitIntent}. &
--- & --- \\
\addlinespace[4pt]
Error rate &
$\displaystyle 1 - \frac{|\mathcal{O}|}{|\mathcal{R}|}$ &
Share of runs that failed. $\mathcal{R}$ is the set of attempted runs. A failed run ends without a final outcome, for example after a second consecutive guard rejection. &
$\downarrow$ & --- \\
\addlinespace[4pt]
Retrieval calls &
$\displaystyle\frac{1}{|\mathcal{O}|}\textstyle\sum_{o\in\mathcal{O}} c(o)$ &
Mean number of knowledge-base lookups per successfully completed run. $c(o)$ counts the calls of \texttt{grep\_knowledge} and \texttt{read\_knowledge\_card} in the run that produced $o$. Flags agents that stop consulting the knowledge base. &
--- & --- \\
\bottomrule
\end{tabular}
\caption{Offline benchmark metrics. $h$ and $g$ are LLM judges with a binary verdict. Both see the dialogue and the agent's final outcome. The hallucination judge $h$ also receives the business rules, tool descriptions, and the tool results obtained by the agent.}
\label{tab:bench-metrics}
\end{table*}

\paragraph{Evaluation on a later sample.}
We used the original 500 conversations to compare reward designs
and shortlist checkpoints (App.~\ref{app:grpo-setup}). We reserved a second sample
of 500 conversations from the same operator-handled stream, logged
two weeks later, for evaluation of the final SFT and GRPO checkpoints.
We excluded conversations and customers present in the training data
or the original benchmark and used the same replay and scoring protocol. 

On the later sample, GRPO returned text answers in 85.4\% of
conversations, compared with 57.4\% for SFT. The hallucination judge
flagged 7.1\% of GRPO text answers and 11.2\% of SFT text answers.
Intent match was 69.3\% for GRPO and 77.2\% for SFT. Results are averaged over three runs.

We report the original benchmark throughout because
all variants were evaluated on it. The consistent findings suggest that the
observed gains are not confined to the benchmark used
for model selection.

\paragraph{Hallucination Rate}
Table~\ref{tab:bench-metrics} summarizes the benchmark metrics. Our primary offline safety measure is the share of customer-facing text answers flagged by a fixed whole-answer hallucination judge. The judge receives the conversation, business rules, tool descriptions, the tool results obtained by the agent, and its final answer. It flags unsupported assertions, distortions or contradictions of the available evidence, and violations of response requirements. The last category includes disclosing information that must be withheld and omitting information that the applicable business rules require the agent to provide. A plausible factual statement can still yield an incorrect consultation when the response breaks a mandatory domain rule. The reported hallucination rate counts answers with at least one flag of
any kind, including rule violations. 
\begin{table}[t]
\centering
\small
\begin{tabular}{@{}lrr@{}}
\toprule
Judge flag & SFT & GRPO (final) \\
\midrule
External knowledge    &  6.0 & 0.8 \\
Context distortion    & 10.2 & 8.0 \\
Context contradiction &  0.2 & 0.2 \\
Policy violation      & 12.2 & 6.3 \\
\midrule
Any flag              & 13.6 & 8.6 \\
\bottomrule
\end{tabular}
\caption{Percentage of \texttt{Answer} outputs receiving each flag
on the offline benchmark. An answer can receive multiple flags.
The last row counts each flagged answer once.}
\label{tab:offline-judge-flags}
\end{table}
We additionally track answers flagged only for response-rule
violations, since these flags expose differences between
the judge's rubric and operational annotation criteria
(App.~\ref{app:judge}). The hallucination judge is applied to \texttt{Answer} outcomes. \texttt{Clarify} outcomes are tracked separately: a follow-up question is not treated as a factual assertion under the human assessment rubric. Unnecessary or inappropriate clarifications are caught by the intent
judge (App.~\ref{app:intent-judge}), but we did not assess the wording of
clarification questions or their suggested replies for unsupported
factual presuppositions.

\paragraph{Complementary Metrics}
A low hallucination rate alone does not make an agent useful: it can be reached by answering less, asking more, or handing off. The remaining rows of Table~\ref{tab:bench-metrics} guard against these shortcuts. \emph{Intent match} checks that the outcome addresses the customer's actual need, whichever outcome type the agent chose. The outcome rates show whether a candidate shifts its answers toward clarifications, handoffs, or exits, and the error rate counts runs that produced no outcome at all. \emph{Retrieval calls} tracks whether the agent keeps consulting the knowledge base, since earlier training runs nearly stopped using it. These dimensions help explain failures but do not replace the whole-answer safety measure. On the original benchmark, we used these measures to compare candidate variants and select one for online evaluation, without applying a single hallucination cutoff across all benchmark iterations. All judges use GPT-5.5 with medium reasoning effort.

\paragraph{Relation to Human Assessment}
The offline judge covers a broader set of factual and rule-following errors than the critical-error category used in human quality assessment. After offline candidate selection, the model enters an A/B test; the product outcomes of that test are reported separately. A successful test is followed by human review of production conversations, where the acceptance target is fewer than 5\% of reviewed segments containing a critical consultation error.

\paragraph{Judge Substitution}
All benchmark judges use GPT-5.5, so candidate selection could depend on
this choice. To test this, we re-scored the text answers of two
candidates, the final SFT checkpoint (App.~\ref{app:sft}) and the
final GRPO checkpoint (App.~\ref{app:grpo-final}), with three other
judge models: GPT-5.2 \citep{gpt52}, and Kimi K3 \citep{kimik3} and DeepSeek-V4-Pro \citep{deepseekv4} from other model families.
Each judge received the same prompt and inputs as the GPT-5.5 judge.

All four judges preserve the ordering of the candidates (Table~\ref{tab:judge-swap}); the difference is significant under each judge (Fisher's exact test, $p<0.05$). Absolute rates differ across judges, so candidates are compared only under the same judge.

\begin{table}[t]
\centering
\small
\setlength{\tabcolsep}{4pt}
\begin{tabular}{@{}lcc@{}}
\toprule
Judge & SFT & GRPO (final) \\
\midrule
GPT-5.5         & 13.6 & 8.6 \\
GPT-5.2         & 14.5 & 9.5 \\
Kimi K3         & 4.5 & 1.5 \\
DeepSeek-V4-Pro & 5.2 & 1.7 \\
\bottomrule
\end{tabular}
\caption{Hallucination rates (\%, lower is better)
of the final SFT and GRPO checkpoints under
different judge models.}
\label{tab:judge-swap}
\end{table}

\section{Judge Calibration}
\label{app:judge_cal}

\subsection{Hallucination Judge}
\label{app:judge}

\paragraph{Evaluation target and human reference.}
Our GPT-5.5 hallucination judge (App.~\ref{app:bench}) flags
unsupported claims, distortions, source contradictions, and violations
of explicit response requirements using the agent's business rules,
retrieved knowledge, and tool results. The dialogue clarifies the
request but is not treated as grounding. The judge evaluates whole
answers separately from the claim-level GRPO reward
(App.~\ref{app:grpo}).

We compared its predictions with consultation-error labels
from the support quality team. Each answer received three reviews,
with up to two additional reviews when annotators disagreed.
These labels are not based on the same grounding evidence:
annotators assess consultation quality against live back-office data
and full customer records, just as support operators do.
The judge, by contrast, evaluates the agent's answer based on
the evidence available to the agent.
Disagreement can therefore reflect different evidence and evaluation
criteria as well as judge errors.

\paragraph{Judge Calibration and Evaluation.}
We calibrated the judge over multiple development rounds, using
disagreements with human annotations to identify and address missing
business rules, improper use of dialogue as grounding evidence,
and data artifacts, including corrupted currency symbols and
name masking. These rounds informed updates to both data
preprocessing and judge instructions. We used 660 historical responses and annotations for development and then fixed the judge
and evaluated it on a held-out benchmark of 1025 answers. The human annotations had already been collected and were
not revised by these preprocessing changes.

We define \emph{gold}($\geq k$) as a binary reference that labels
an answer positive if at least $k$ annotators flag a consultation
error. During development, we prioritized recall, accepting
additional flags for review to reduce missed errors.
Consistent with this objective, we use \emph{gold}($\geq1$)
as our primary reference, because it retains every answer flagged by at least one annotator rather than restricting evaluation to cases with broader
annotator agreement. This choice also includes disputed labels:
in the development sets, 26--30\% of its positives were flagged
by only one annotator.

On the held-out benchmark, the final judge achieved precision
of 0.83, recall of 0.74, and $F_1$ of 0.782 against this reference. We also report Cohen's $\kappa$
as 0.378 alongside these metrics to account for chance agreement given
the marginal label frequencies \cite{rao2026agreement}.

\paragraph{Disagreement analysis.}
We manually reviewed all 291 disagreements with the
\emph{gold}($\geq1$) reference. The audit identified 40 answers
in which the judge correctly detected omissions of required
information that were not flagged by the human annotators,
and 70 answers in which human error flags arose from currency
symbols corrupted by the annotation interface. These corruptions
were absent from the original agent responses. Together, these
two categories accounted for 110 disagreements (37.8\%).

A further 67 disagreements concerned additions that were
unsupported by the agent's evidence but considered benign
under the consultation-quality criteria. Such additions fall
within the judge's stated grounding rubric even when they do
not materially impair the consultation. Thus, 177 of the 291
disagreements (60.8\%) reflected annotation artifacts or differences
in evaluation criteria. The audit also identified 103 substantive
errors missed by the judge and 11 other disagreements.

We additionally reviewed all 124 jointly
positive answers for which the judge reported only
response-requirement violations. In 69 cases,
the audit confirmed valid violations missed by
annotators, whose flags instead concerned
interface-induced currency corruption.
Annotators identified the same issue in 32 cases, a different
error in 18, and an unclear issue in five. Across both audited
groups, currency artifacts accounted for human flags in
139 of 1025 answers (13.6\%).

Table~\ref{tab:judge-counterfactual} reports agreement after
reconciling manually identified interface artifacts and
required-information omissions with the judge's evaluation
rubric. For \emph{gold}($\geq1$), this reconciliation changes
110 reference labels and increases Cohen's $\kappa$ from
0.378 to 0.619, without changing the judge's predictions. For the 69 jointly positive cases, confirmed
response-requirement violations replace currency artifacts
as the basis for the positive reference labels,
which therefore remain unchanged.

The manual audit provides evidence that the judge identifies grounding and response-compliance violations that the operational annotations systematically miss, while annotation-interface artifacts account for a substantial fraction of the apparent judge errors.

\begin{table}[t]
\centering
\small
\setlength{\tabcolsep}{7pt}
\begin{tabular}{@{}lrr@{}}
\toprule
Human reference & Raw $\kappa$ & Counterfactual $\kappa$ \\
\midrule
gold($\geq1$) & 0.378 & 0.619 \\
gold($\geq2$) & 0.405 & 0.644 \\
gold($\geq3$) & 0.434 & 0.671 \\
\bottomrule
\end{tabular}
\caption{Agreement with the original human reference and after
reconciliation of manually identified annotation artifacts
and rubric discrepancies.}
\label{tab:judge-counterfactual}
\end{table}

\paragraph{Implications for evaluation.}
Raw agreement with consultation labels understates
the judge's demonstrated detection capabilities: confirmed
grounding and response-compliance violations can be missed
by annotators, while interface artifacts introduce spurious
human error flags. For screening, we therefore prioritize
recall over reproducing consultation-error labels, accepting
review of benign unsupported additions. For relative model
comparison, all four judges preserve the ordering of the SFT
baseline and the final GRPO candidate despite differing
absolute flag rates (App.~\ref{app:bench}), showing that
this ordering is robust to the choice among the tested judges.

\subsection{Intent Match Judge}
\label{app:intent-judge}

\paragraph{Task and human reference.}
The intent match judge based on GPT-5.5 evaluates whether an agent's response
addresses the customer's request and follows the required
interaction behavior. It flags unrelated answers, unnecessary
or inappropriate questions, ignored handoff requests, and
failure to acknowledge negative sentiment. Factual grounding
is evaluated separately (App.~\ref{app:judge}). Calibration treats a behavioral error as the positive class:
an answer receives a positive reference label if at least
three of five annotators identify the same behavioral-error
category. We used 276 responses for development and evaluated transfer
on a held-out cohort of 270 responses from other agent
variants, annotated by the same human assessment group.
Neither automatic nor manual prompt refinement used this
held-out cohort.

\paragraph{Error-driven prompt refinement.}
We refined the judge through an iterative procedure related
to textual-gradient prompt optimization
\cite{pryzant2023automaticpromptoptimizationgradient} and calibration of
LLM evaluators against human judgments
\cite{liu2023calibratingllmbasedevaluator}. At each iteration, we evaluated the current prompt on all
development responses and sampled false positives and false
negatives. An LLM editor received the current instructions,
sampled dialogues, judge verdicts and rationales, and human
votes and comments. It diagnosed a recurring failure pattern
and proposed a targeted rewrite of the editable instruction
block, preserving the error taxonomy and output schema. We retained a revision only if it improved error-detection
$F_1$ on the full development cohort.
Rejected revisions left the current best prompt unchanged, and
the next iteration used a fresh error sample. The search
terminated at its budget limit or after repeated rejected
revisions. Manual refinement, also restricted to development
data, produced the final prompt. The revisions clarified concrete decision boundaries:
repeated status explanations should be flagged when they
ignore a customer's objection or new information; redundant
questions fall under inappropriate clarification; and an
apparently unanswered handoff request should not be penalized
when transfer has already occurred.

\paragraph{Calibration results.}
The revised prompt improved
error detection beyond the development responses
(Table~\ref{tab:intent-judge-calibration}). 
The largest gain was in recall, accompanied by higher
precision, indicating that the revised instructions
corrected systematic omissions rather than merely
encouraging more frequent error flags.

\begin{table}[t]
\centering
\small
\setlength{\tabcolsep}{6pt}
\begin{tabular}{@{}llrrrr@{}}
\toprule
Split & Prompt & $P$ & $R$ & $F_1$ & $\kappa$ \\
\midrule
Development & Baseline & 0.75 & 0.38 & 0.50 & 0.37 \\
            & Revised  & 0.81 & 0.70 & 0.75 & 0.65 \\
\midrule
Held-out    & Baseline & 0.61 & 0.42 & 0.50 & 0.37 \\
            & Revised  & 0.65 & 0.75 & 0.70 & 0.60 \\
\bottomrule
\end{tabular}
\caption{Intent-judge performance against majority human
behavioral-error labels.
$P$, $R$, and $F_1$ denote precision, recall, and $F_1$ for behavioral errors.}
\label{tab:intent-judge-calibration}
\end{table}

\section{Reward Design and GRPO Training}
\label{app:grpo}

Hallucination assessment checks claims against the
agent's available evidence and answers against explicit
response requirements. Helpfulness is
much harder to annotate consistently, since several different answers
can be acceptable and pairwise preference labels are costly and noisy.
Yet helpfulness directly affects automation: a correct but incomplete
answer still sends the customer to an operator. We therefore treated
helpfulness as an open measurement problem and tried several reward
designs, each relying on a different helpfulness proxy: coverage of
supported operator claims, relative judging, rulebook gate,
operator response style, and customer intent.

\subsection{What to Reward: Precision and Coverage of Verified Claims}
\label{app:grpo-what}

Our training setup uses real customer dialogues collected in production, together
with the database records and knowledge documents needed to answer the
model's tool calls. We compared reward designs and shortlisted checkpoints on the
original 500-conversation benchmark, which was disjoint from
the training data. 

Each training example pairs a customer dialogue with the historical human
operator reply, the knowledge base, and the dialogue's recorded tool
data. Each reference operator reply is split into a set $S$ of separate 
statements (claims) as discussed in App.~\ref{app:kb}. 
Some unimportant parts of the human's reply, such as greetings, are discarded. 
Real human answers may also contain information that is not grounded in the 
data available to the LLM agent. It can include prior human knowledge 
or information from web tools that are not available to the agent.
Therefore, we classify all the meaningful operator claims as \texttt{supported} 
if they are grounded in the information available to the LLM agent 
from knowledge base or tool calls.

We score the agent's answer using two variants
of coverage and grounding precision. 
Coverage $c_{\mathrm{all}}$ assesses how fully the agent's answer
expresses the meaningful claims in the operator reply, using all claims
in $S$ regardless of their support labels. It rewards completeness relative
to the human response. Supported coverage $c_{\mathrm{supp}}$ assesses completeness
against only the claims in $S$ labeled \texttt{supported}. Filtering the
reference removes claims for which the verification procedure found
insufficient evidence. Precision $p$ assesses how well the factual claims in the agent's
answer are supported by the evidence available to it, including retrieved
knowledge and tool results. Note that precision does not rely on the 
recorded human operator claims, and is judged over the actual evidence
retrieved by the agent.

We use RULER (Relative Universal LLM-Elicited Rewards), a group-relative
judging method for GRPO \citep{openpipe2025ruler}, to score claim-level
grounding and coverage. The judge compares all sampled trajectories for the
same dialogue, yielding precision $p$ and coverage $c$ for each response.

A natural choice to combine precision
and coverage (recall) is $F_1$-metric
\begin{equation}
F_1(p, c) = \frac{2pc}{p + c},
\end{equation}
where we set $F_1(0,0) = 0$. 

We compare several simple reward designs in Table~\ref{tab:grpo-what}.
Hallucination rate is the share of text answers flagged by the whole-answer judge for any reason: an unsupported assertion, a distortion or contradiction of the evidence, or a violation of a response requirement, including rule violations (Appendices~\ref{app:bench} and~\ref{app:judge}).
Intent match is the share of evaluated outcomes judged to address the
customer's need by the benchmark's intent-alignment judge
(App.~\ref{app:bench}).
Text-answer rate is the share of final outcomes returned as
\texttt{Answer}. The grounding-and-coverage reward applies to
\texttt{Answer} and nonempty \texttt{Clarify}
responses. Responses judged to contain no factual
claims, including pure clarification questions,
have their precision score capped at 0.6. Non-text outcomes use separate heuristic eligibility
checks. \texttt{ExitIntent} and \texttt{Handoff}
are eligible when the latest customer message matches
a human-operator keyword pattern, or when the retrieval
target is empty and no rollout in the sampled group
calls a customer-data tool. Empty \texttt{Clarify}
outcomes are eligible only under the latter condition.
Ineligible outcomes receive zero reward. Under the final reward, eligible non-text outcomes
receive less credit when the same rollout group
contains a higher-scoring textual response.
Trajectory penalties also apply to these outcomes.

\begin{table}[t]
\centering
\small
\setlength{\tabcolsep}{3pt}
\begin{tabular}{@{}lrrr@{}}
\toprule
Reward & Halluc.\,$\downarrow$ & Intent\,$\uparrow$ & Answer \\
\midrule
Precision $p$ & 24.8 & 61.8 & 77.6 \\
Coverage $c_{\mathrm{supp}}$ & 76.7 & 86.8 & 81.2 \\
Coverage $c_{\mathrm{all}}$ & 82.4 & 89.4 & 81.6 \\
$F_1(p,c_{\mathrm{supp}})$ & 32.5 & 68.3 & 81.2 \\
$F_1(p,c_{\mathrm{all}})$ & 39.1 & 72.6 & 81.2 \\
\bottomrule
\end{tabular}
\caption{Precision and coverage reward designs.
Hallucination rate (Halluc.), intent match (Intent),
and text-answer rate (Answer) are percentages.
Coverage targets include either all operator claims
or only verified claims.}
\label{tab:grpo-what}
\end{table}

Precision alone encourages conservative answers: the agent can avoid
unsupported assertions by giving a short reply, leaving parts of the
request unanswered, or deferring to a human operator. This reward has no
explicit incentive to include missing information. Its lower hallucination
rate comes with lower intent match and a lower text-answer rate than the
coverage-based variants, consistent with a more evasive response strategy.

Coverage alone rewards reproducing operator facts without penalizing
unsupported additions, yielding high intent match but substantially more
hallucinations than precision alone. Restricting the target to supported
operator claims reduces incentives to repeat facts unavailable to the
agent and lowers hallucinations.
Based on this comparison, we retain verified operator
claims as coverage targets in all subsequent reward
designs and write $c$ for $c_{\mathrm{supp}}$.
Combining coverage with precision in $F_1(p,c)$ balances substantive
answers with factual grounding: compared with precision alone, it improves
intent match and increases the text-answer rate, but still raises
hallucinations.

\subsection{Retrieval Component}
\label{app:grpo-judge}

Earlier runs with $F_1$ nearly stopped using the knowledge-base tools, allowing the
agent to answer from memorized patterns without obtaining the relevant
evidence and increasing hallucinations. The retrieval component of the reward was introduced to counter this behavior:
it encourages the agent to consult the required sections before answering,
while retaining precision as the check on how that evidence is used.

For each dialogue with an annotated retrieval need,
$q$ is the fraction of the relevant knowledge-base
sections that the agent retrieved during its trajectory.
When no retrieval is required, $q$ equals zero.
The reward adds $r_{\mathrm{ret}} = 0.2pq$ to the
$F_1$ core and caps the result at one.
Unlike standalone agentic retrievers optimized for
evidence recall \citep{tsymboi2026t}, we scale retrieval
credit by answer precision $p$, reducing the reward
when the answer contains unsupported assertions.

\begin{equation}
R_{\mathrm{rec}} = \min(1,F_1(p,c)+r_{\mathrm{ret}}).
\label{eq:grpo-rec}
\end{equation}

The component increased retrieval of relevant sections
and modestly reduced hallucinations
(Table~\ref{tab:grpo-comparison}).
Increasing its coefficient from 0.2 to 0.4 instead
raised the hallucination rate to 30.7\%. A larger weight allows the reward to reach its cap
at lower grounding and coverage scores, weakening
the incentive to improve these once the cap is reached.
This reduced incentive may contribute to the
higher hallucination rate. We therefore retained the weight
of 0.2 to discourage skipping required retrieval
without allowing retrieval credit to dominate
the grounding and coverage objective.

\begin{table}[t]
\centering\small
\setlength{\tabcolsep}{3pt}
\begin{tabular}{@{}lrrr@{}}
\toprule
Variant & Halluc.\,$\downarrow$ & Intent\,$\uparrow$ & Answer \\
\midrule
$F_1(p,c)$
    & 32.5 & 68.3 & 81.2 \\
Retrieval component~(\ref{eq:grpo-rec})
    & 26.7 & 60.2 & 70.0 \\
Rulebook gate~(\ref{eq:grpo-policy})
    & 17.2 & 59.1 & 79.8 \\
Style reward~(\ref{eq:grpo-si})
    & 23.5 & 72.4 & 78.2 \\
    $p$-dominant core + gate  & 12.3 & 60.5 & 78.2 \\
Weighted rubrics~(\ref{eq:grpo-think})
    & 8.1 & 70.5 & 89.3 \\
\midrule
\textbf{Final}~(\ref{eq:grpo-strict})
    & 8.6 & 71.5 & 88.9 \\
\bottomrule
\end{tabular}
\caption{Offline comparison of reward designs.
Hallucination rate (Halluc.), intent match (Intent),
and text-answer rate (Answer) are percentages.
The bold label marks the final reward.}
\label{tab:grpo-comparison}
\end{table}

\subsection{Rulebook Gate}
\label{app:grpo-constraints}

Grounding checks whether a statement is supported by evidence, but a
supported statement can still be inappropriate to disclose or insufficient
for a correct consultation. For example, a tool may expose the inviter's
bonus while the customer is the invited friend; repeating that amount can
be factually grounded and still violate the disclosure rules. We therefore
add a separate compliance assessment against a rulebook assembled from the
domain rules, tool-field instructions, and response-format instructions
already included in the agent's system prompt. It covers disclosure restrictions,
participant roles, and the handling of customers ineligible for the program.

The compliance judge receives the rulebook, customer dialogue, case
records, and each candidate's retrieved material, without the historical
operator reply. It compares the sampled answers for violations of explicit
rules, regardless of whether their statements are factually supported. If
none violates a rule, all receive full compliance credit. The judge does
not favor refusal over a substantive answer that follows the rules. Its
score $g$ is higher for better adherence. We apply it as a multiplicative
gate:
\begin{equation}
\begin{aligned}
 R_{\mathrm{pol}}
 &=\min\!\left(1,F_1(p,c)+r_{\mathrm{ret}}\right)\\
 &\qquad{}\times\bigl(0.4+0.6g\bigr).
\end{aligned}
\label{eq:grpo-policy}
\end{equation}
The gate penalizes rule violations even when grounding
and coverage are high, while retaining an empirically
chosen minimum of 40\% of the pre-gate reward.

Compared with the retrieval-bonus baseline, the policy reward lowers
the hallucination rate and increases the text-answer rate, while
intent match decreases (Table~\ref{tab:grpo-comparison}). The evaluation judge counts rule violations for every
variant, so the comparison uses the same rubric and
isolates the effect of adding the compliance gate.

\subsection{Helpfulness Proxies: Operator Style Cloning}
\label{app:grpo-proxies}

We also explored operator-style imitation as a proxy for helpfulness.
Direct comparison with original operator replies introduced two
problems. First, judges continued to penalize differences in
case-specific facts even when instructed to disregard them.
Second, some operator replies relied on information unavailable
in the agent's knowledge base or tools, making their content
infeasible to reproduce.

Customer dialogues were clustered by customer intent and
operator-response topic into 15 scenario classes, each associated
with an answer template. Each template provides an ideal answer
sample and the expected sequence of response blocks for its scenario.
Case-specific values, such as bonus amounts and deadlines, were
replaced by placeholders. For template-based comparisons, we selected
an applicable template whose required content could be supported
by the agent's sources.

We classified every training dialogue against these template classes.
The style judge compares sampled answers with the template for their
scenario, or with the historical operator reply when no scenario is
assigned. Its score $s$ reflects whether an answer follows the expected
sequence of logical steps, level of detail, and tone.
The judge is instructed not to assess whether customer-specific facts
are correct; the grounding reward assesses them separately.

Style alone does not ensure that the response resolves the current
request. We therefore combine it with feedback from the same intent judge
used to measure intent match in Table~\ref{tab:grpo-comparison}.
The judge reads the customer dialogue and candidate answer, checking
whether the response addresses the customer's need and whether
clarification or human handoff is appropriate. Its score $i$ brings this
evaluation criterion directly into training. The selected blend replaces
coverage inside the harmonic mean while retaining grounding, retrieval,
and the policy gate:
\begin{equation}
\begin{aligned}
 R_{\mathrm{sty}}
 &=\min\!\bigl(1,F_1(p,0.7s+0.3i)+r_{\mathrm{ret}}\bigr)\\
 &\qquad{}\times\bigl(0.4+0.6g\bigr).
\end{aligned}
\label{eq:grpo-si}
\end{equation}
This combines the two aims: operator-like response structure and attention
to the customer's particular need. 

Compared with the policy reward, the style/intent blend improves
intent match, but increases hallucinations and slightly lowers
the text-answer rate (Table~\ref{tab:grpo-comparison}). 

\paragraph{Evaluation of template conformity.}
We evaluated template conformity on a separate randomly sampled
held-out set of conversations, distinct from the 500-conversation
offline benchmark. We used the same checkpoints and inference
settings as in the online experiment.

We used two evaluation protocols. In template-guided pairwise
evaluation, the judge compared the two candidate answers against
the applicable template and selected the answer that better
satisfied its content and presentation requirements.
The judge was GPT-5.5. It received both candidate answers, the scenario
template, and an ideal answer example with substituted values.
Each answer pair was evaluated in both presentation orders
(A/B and B/A), with each judgment counted as a separate vote.
We report the win rate as $W/(W+L)$, excluding ties, where $W$
and $L$ count votes favoring the style-reward and final-reward
GRPO answers, respectively.

The pointwise \texttt{style\_success} measure assigned each textual
answer a graded score in $[0,1]$, reflecting sentence-level
correspondence with the template and stylistic alignment with
the ideal answer example. For each checkpoint, scores were averaged
over 500 retained textual answers. Non-text outcomes were assessed
using intent match and excluded from these averages.

The paired pointwise comparison used only the intersection of
conversations with retained textual answers from both checkpoints
(476 matched conversations). Its confidence interval was
estimated by bootstrapping these matched conversations.

When no applicable template was available, the historical operator reply
served as the reference; these cases accounted for less than 3\% of the
evaluation sample. The pairwise win rate was computed from 415 non-tied
votes, counting each A/B and B/A judgment separately.

The style-reward checkpoint achieved a pairwise win rate of 74.2\%
against final-reward GRPO (95\% CI: 69.9--78.5\%) and a higher
pointwise \texttt{style\_success} score
(0.494 vs.\ 0.389; Table~\ref{tab:style-proxies}).
The paired pointwise difference was +10.5 percentage points
(95\% CI: 3.7--19.1). Confidence intervals were estimated
by bootstrapping at the conversation level.

\paragraph{Human calibration.}
We calibrated the pairwise judge against human annotations on
1500 randomly sampled answer pairs. Three annotators evaluated
each pair, selecting the better answer, and majority voting
produced the human reference. The calibration data were semantically deduplicated and
decontaminated against both the offline benchmark and the
template-conformity evaluation set. Agreement between the judge and the human
reference was Cohen's $\kappa = 0.63$.

The style reward improved template conformity,
but this gain did not translate into higher online
automation (App.~\ref{app:online}).
Template conformity was therefore insufficient
as a proxy for operational usefulness.

\begin{table}[t]
\centering
\small
\begin{tabular}{@{}lcc@{}}
\toprule
Model & Pairwise win rate & \texttt{style\_success} \\
\midrule
GRPO (final) & Reference & 0.389 \\
GRPO (style) & 74.2\% & 0.494 \\
\bottomrule
\end{tabular}
\caption{Template-conformity evaluation on a separate held-out sample.
Pointwise means use 500 textual answers per checkpoint.
Pairwise wins are measured against GRPO (final) using
415 non-tied votes across both answer orders.
The paired pointwise comparison uses the intersection of conversations
with retained textual answers from both checkpoints.
Confidence intervals are reported in the text.}
\label{tab:style-proxies}
\end{table}

\subsection{Rubrics from Product Feedback}
\label{app:rubrics}

The rubric categories behind the violation severity $v$
(App.~\ref{app:grpo-final}) were derived from reviewer feedback rather
than designed in advance. Support-quality assessors reviewed 674 agent answers sampled from offline runs on production conversations that
the existing bot had not automated, excluding
conversations overlapping with the benchmark.
They left free-text comments describing problems
with the answers. We grouped the comments by error type using an LLM and manual review.
We excluded subjective preferences, such as disagreements over greetings
or phrasing, and formulated 14 candidate checks for recurring errors.
Each check produced a binary violation flag, either deterministically
or through an LLM judge.

For each candidate check, we estimated dialogue-level flag rates
for responses from the GRPO checkpoint trained with the rulebook
gate (reward $R_{\mathrm{pol}}$) and for historical operator replies. We compared the two sets of replies on the same conversations used
for expert review.
We computed a 95\% bootstrap confidence interval for each rate.
We retained a check only if its flag rate on policy responses
exceeded 5\% and the lower bound of the policy interval exceeded
the upper bound of the operator interval. Ten of the 14 candidate
checks met both criteria.

We group the ten retained checks into three categories:
\begin{itemize}
\item \textbf{Rule compliance, verification, and privacy (3 checks).}
These checks cover mandatory support rules, required lookups of
customer records, and restrictions on disclosing bonus information.
\item \textbf{Bonus explanations and next steps (4 checks).}
These checks assess whether the agent explains the customer's bonus
situation and gives the appropriate next step, for example when an
expected payment is delayed.
\item \textbf{Relevance and delivery of unfavorable outcomes (3 checks).}
These checks flag misleading extra detail and inappropriate ways of
communicating an unfavorable decision.
\end{itemize}

We manually assigned each retained check a fixed weight between 0.1
and 0.5. We considered its overall frequency, the difference in flag
rates between the rulebook policy and operators, the stability of
the flag across repeated generations from the same model, and its
association with human acceptability labels. Violations of mandatory rules and restricted disclosures received the highest weight (0.5). Intent-miss severity is scored separately from these ten binary
checks. App.~\ref{app:grpo-final} describes how it affects
coverage and the violation penalty.
 
\subsection{Final Reward}
\label{app:grpo-final}

The preceding
rewards still left substantial hallucination rates in offline evaluation,
so we sought stronger pressure for factual grounding before selecting a
production candidate. In the harmonic mean, low coverage also weakens
the incentive to improve precision, even when the answer contains
unsupported claims. We therefore replace the harmonic mean with a
precision multiplier and a coverage-dependent contribution. We denote the violation penalty factor by
$w_{\mathrm{rule}} = 0.4 + 0.6(1-v)$.
The resulting reward is
\begin{equation}
R_{\mathrm{wgt}}
= \min(1, p(0.3+0.7c_{\mathrm{eff}}) + r_{\mathrm{ret}})
  \,w_{\mathrm{rule}}.
\label{eq:grpo-think}
\end{equation}
The core awards $0.3p$ at zero effective coverage and increases
to $p$ at full effective coverage.
The retrieval bonus retains its dependence
on precision.

The violation severity $v$ broadens the earlier rulebook assessment to
include recurring support failures. The rubric covers missing required
tool calls, excessive elaboration, inappropriate delivery of unfavorable outcomes, misleading explanations of delays or bonus changes, and
privacy violations. The rubric categories and their construction are described in App.~\ref{app:rubrics}. Unlike $g$, which is a
holistic compliance score assigned by a judge, $v$ is computed explicitly
by aggregating weighted violation flags. The judge assigns binary
flags for the rubric categories, supplemented by deterministic privacy
checks. Let $b$ be the sum of the weights of active rubric flags and
$s_{\mathrm{int}}$ the separately assessed intent-miss severity.
For a textual response with at least one verified operator claim,
we set $v=\min(1,b+0.15s_{\mathrm{int}})$. We normally cap coverage
at intent fulfillment, setting
$c_{\mathrm{eff}}=\min(c_{\mathrm{supp}},1-s_{\mathrm{int}})$.
If the refusal heuristic identifies an honest refusal on an
unanswerable request, we instead set
$c_{\mathrm{eff}}=\max(c_{\mathrm{supp}},0.5)$. This exception
lifts the intent cap on coverage while retaining the intent penalty
in $v$.

When there are no verified operator claims, intent fulfillment
determines coverage directly: $c_{\mathrm{eff}}=1-s_{\mathrm{int}}$.
In this case, only rubric violations contribute to
$v=\min(1,b)$.
The factor $w_{\mathrm{rule}}$ gives less credit to
answers with more severe violations while retaining the same floor as
the earlier policy gate.

With the retrieval bonus and rulebook gate held fixed, replacing the
\(F_1\) core with the precision-dominant core reduced the hallucination
rate from 17.2\% to 12.3\%. Keeping this new core fixed and replacing
the rulebook gate with weighted rubrics further reduced the hallucination
rate to 8.1\%, while intent match rose from 60.5\% to 70.5\% and the
text-answer rate rose from 78.2\% to 89.3\%. This comparison measures
the combined effect of the rubric changes; the contribution of each
individual rubric remains unknown.

The final variant also penalizes costly or malformed agent trajectories:
\begin{equation}
 R=R_{\mathrm{wgt}}\,w_{\mathrm{traj}}.
\label{eq:grpo-strict}
\end{equation}
Each of three conditions halves the trajectory factor $w_{\mathrm{traj}}$: a model
call exceeding 5K generated tokens, more than 32 consecutive whitespace
tokens, or a textual ending without the required final-result tool call.
A trajectory budget further multiplies the penalty by $\min(1,(5000/T)^2)$,
where $T$ counts generated tokens across all calls in the trajectory.
The token budget of 5K was set at the 99th
percentile of the GPT-5.5 agent's trajectory length
to discourage unusually long generations.
These penalties discourage prolonged generation and malformed endings
without truncating an individual verification step solely for exceeding
the soft token budget.

The two variants performed similarly offline (Table~\ref{tab:grpo-comparison}), with
differences of at most one p.p. across all three metrics.
We retained the trajectory penalty for production to discourage
excessively long generations and malformed outputs.

\begin{table}[t]
\centering
\small
\setlength{\tabcolsep}{4pt}
\begin{tabular}{@{}lrrr@{}}
\toprule
Reasoning budget & Halluc.\,$\downarrow$ & Intent\,$\uparrow$ & p99 tokens \\
\midrule
Disabled      & 15.3 & 58.3 & 1081 \\
1K tokens & 12.7 & 74.1 & 1521 \\
3K tokens & 10.3 & 73.1 & 1931 \\
\textbf{5K tokens} &  8.6 & 71.5 & 1937 \\
Unlimited &  8.4 & 73.5 & 2684 \\
\bottomrule
\end{tabular}
\caption{Inference-time reasoning budgets for the final GRPO agent.
Hallucination rate and intent match are percentages;
p99 tokens is the 99th percentile of tokens generated per
conversation. The bold label marks the deployed 5K-token setting.}
\label{tab:reasoning}
\end{table}

\paragraph{Reasoning budget at inference.}
To choose a serving budget, we held the final GRPO checkpoint fixed
and varied its reasoning-token limit (Table~\ref{tab:reasoning}).
Allowing 5K reasoning tokens reduced the hallucination rate from
15.3\% with reasoning disabled to 8.6\%. Removing the cap reduced
hallucinations by a further 0.2 percentage points and raised intent
match by 2.0 points, but increased the 99th percentile of total
generated tokens per conversation from 1937 to 2684. We therefore
retained the 5K reasoning cap for serving.

\subsection{Effect of SFT Initialization}
\label{app:grpo-sft}

Our main GRPO experiments start from the off-the-shelf model.
We separately investigated supervised fine-tuning using
rewritten operator replies and teacher-generated trajectories
(App.~\ref{app:sft}). Here, we evaluate whether a
teacher-distilled checkpoint provides a better initialization
for GRPO.

We use the AugExit (5\%) + hallucination filtering checkpoint (Table~\ref{tab:rf-v3-answer-types}),
obtained by full-parameter training on filtered teacher-generated support trajectories.
We then apply GRPO using the training recipe described in
App.~\ref{app:grpo-setup}.

We use the final reward $R$
(Eq.~\ref{eq:grpo-strict}), the precision-dominant score with the retrieval bonus and the violation-severity gate, multiplied by the trajectory factor $w_{\mathrm{traj}}$. Keeping the reward fixed lets us isolate the effect of initialization.

\begin{table}[t]
\centering\small
\setlength{\tabcolsep}{3pt}
\begin{tabular}{@{}llrrr@{}}
\toprule
Experiment & Halluc.\,$\downarrow$ & Intent\,$\uparrow$ & Answer \\
\midrule
SFT  & 13.6 & 80.4 & 60.2 \\
\textbf{GRPO (final)} & 8.6 & 71.5 & 88.9 \\
SFT$\to$GRPO & 9.4 & 74.2 & 77.0 \\
\bottomrule
\end{tabular}
\caption{SFT and GRPO policies with different
initializations. Both GRPO variants use the final
reward $R$. Hallucination rate (Halluc.), intent match
(Intent), and text-answer rate (Answer) are percentages.
The bold label marks the deployed checkpoint.}
\label{tab:grpo-sft}
\end{table}

The teacher provides useful supervision for tool use and response
structure, but its answers still contain hallucinations. Both GRPO
policies achieve lower hallucination rates than the teacher and the SFT
baseline (Tables~\ref{tab:rf-v3-answer-types} and~\ref{tab:grpo-sft}).
Starting GRPO from the off-the-shelf model also yields a higher text-answer
rate than SFT initialization, with lower intent match. We
hypothesize that SFT transfers some of the teacher's response and routing
preferences, which subsequent GRPO only partly revises. Direct reward
optimization allows the policy to explore alternative strategies for
producing grounded answers. SFT therefore remains a useful training
approach, but may not provide the best initialization for our objective of
increasing automation while controlling hallucinations.

\subsection{Training Setup}
\label{app:grpo-setup}
Table~\ref{tab:grpo-setup} gives the configuration used for our GRPO reward experiments. In all ablations except for the SFT-initialized variant in 
App.~\ref{app:grpo-sft} we initialize the policy from
off-the-shelf Qwen3.6-27B and use GPT-5.5 for
training-time judgments.

We construct the retrieval-hard training set from a pool of 10000
training dialogues annotated for claim support and retrieval need.
For each supported operator claim, a judge identifies whether its
evidence is already available in the recorded tool results,
or always-present rules, or instead requires a retrievable KB document.
For the latter claims, a quoted passage is matched to its KB section to
form the retrieval target. We retain all 2135 retrieval-dependent
dialogues with supported claims, including cases requiring multiple
sections. We then randomly sample 2135 dialogues
that do not require KB retrieval. This produces 4270
training examples with equal representation of the two groups, keeping
retrieval necessary without making a KB call useful on every example.

For $N$ sampled responses to dialogue $x$, we compute the advantage of
response $j$ by centering its final scalar reward within the group:
\begin{equation}
 A_{xj}=R_{xj}-\frac{1}{N}\sum_{k=1}^{N}R_{xk}.
\label{eq:grpo-advantage}
\end{equation}
Here $R_{xj}$ includes the applicable reward gates and trajectory
penalties. We omit division by the group's reward standard deviation,
following the mean-centered advantage used in Dr.~GRPO
\citep{liu2025understanding}. This avoids giving greater weight to a
prompt solely because its sampled rewards have a smaller spread.
Groups with identical rewards supply no reward-based advantage.

To select checkpoints for the final production tests, which assess
automation directly, we combined the offline benchmark described in
App.~\ref{app:bench} with human domain assessment.
We first shortlisted the strongest checkpoints using
hallucinations as the primary grounding metric, alongside
intent match and text-answer rate. Human domain specialists then
reviewed the agent's responses on 20 dialogues sampled from the
benchmark pool. They assessed whether the agent behaved appropriately
and checked for unintended, erratic, incoherent, or harmful behavior.
This review provided an assessment of response style and policy
compliance independent of the automatic judges. Together, the offline
metrics and human review informed candidate selection for the final
production tests, where automation was measured directly.

\begin{table}[t]
\centering\footnotesize
\begin{tabular}{@{}p{3.1cm}p{3.6cm}@{}}
\toprule
Parameter & Value \\
\midrule
Policy initialization & Qwen3.6-27B \\
Training examples & 4270 \\
LoRA rank & 128 \\
LoRA $\alpha$ & 256 \\
GRPO steps & 40 \\
Prompts per training step & 256 \\
Rollouts per prompt & 8 \\
PPO epochs & 1 \\
PPO minibatch & 16 \\
Learning rate & $5\times10^{-5}$ \\
Weight decay & 0.1 \\
LR warmup & 10 steps \\
LR decay & 40 steps \\
KL loss & Off \\
Group std normalization & Off \\
Prompt token limit & 6636 \\
Response token limit & 12K \\
Agent rollout & Up to 16 assistant turns \\
\bottomrule
\end{tabular}
\caption{GRPO training configuration.}
\label{tab:grpo-setup}
\end{table}

\section{Supervised Fine-Tuning}
\label{app:sft}

We perform full-parameter supervised fine-tuning (SFT) for tool-based customer support using two sources of supervision: historical operator replies adapted to the agent's available evidence, and complete agent trajectories generated by a teacher.

\subsection{SFT on Operator Replies}
\label{app:sft-operator-targets}
Operator replies offer examples of useful content and presentation but may
rely on information unavailable to the agent. As described in App.~\ref{app:agent}, the agent retrieves information through a fixed set of tools and cannot perform servicing actions. Human operators can consult additional internal context and carry out actions beyond this interface. Their replies may therefore combine general product rules with customer-specific facts and procedural knowledge unavailable to the agent. In the revised-KB run (App.~\ref{app:kb}), Walker found full support for 74.8\% of operator factual claims.

To address this mismatch, we rewrote operator replies to remove unsupported claims while retaining useful content and the operator's style. An offline annotator checked claims against the full knowledge base and customer records. The rewriter received these annotations and supporting excerpts, the operator reply, recorded tool results, and basic domain rules. A rewritten reply could replace the agent's final answer only when the original trajectory ended with \texttt{Answer} (70.5\% of samples). The dataset included the original handoff and exit trajectories so that these actions remained represented.

We performed full-parameter fine-tuning of Qwen3.6-27B
on this dataset and obtained an intent match rate of 55.2\%, significantly lower than that of the GPT-5.5 teacher baseline (Table~\ref{tab:rf-v3-answer-types}).

A limitation of this construction is that grounding checks assess support for individual claims, while the usefulness of the rewritten answer requires a separate assessment, which we did not conduct. Improving evidence support through rewriting does not by itself ensure preservation of the answer's original intent \cite{gao2023rarrresearchingrevisinglanguage}. Operator replies draw on broader customer context, and claims supported by that context may carry information essential to resolving the request. Removing those claims can therefore leave a grounded but incomplete answer.

Preserving usefulness may require reformulating the response around what the agent can establish, including retrieving additional evidence or choosing a different terminal action. Such adaptation approaches the task of generating a new agent trajectory. We therefore chose teacher-generated trajectories for subsequent SFT experiments, aiming to produce targets that both address the customer's need and remain realizable from the agent's available information.

\subsection{Teacher Trajectory Collection and Filtering}

We construct a dataset by running a GPT-5.5 teacher on real support conversations using logs collected from production.

For text responses and clarification requests, the terminal action schema enforces a three-stage generation procedure: \emph{Draft $\rightarrow$ Self-Critique $\rightarrow$ Final}. Since the teacher's internal reasoning traces are not accessible, we introduce this structured procedure to elicit an explicit, observable revision process. The self-critique stage prompts the teacher to verify its draft against retrieved evidence, identify unsupported claims and overlooked constraints, and revise the response before finalization.

To improve supervision quality, we apply two complementary filtering procedures.

First, we employ a hallucination judge
(App.~\ref{app:judge}) to evaluate complete
agent-generated textual responses against the available
grounding context. Trajectories containing flagged
responses are excluded from the corresponding filtered
training dataset. Non-textual terminal actions, for which
hallucination assessment is not applicable, are retained.

Second, we exclude episodes that do not contain a terminal action. Such trajectories represent incomplete agent interactions and lack a final supervision target. This criterion also filters out excessively long trajectories that fail to reach a terminal action within the execution limits.

The resulting dataset contains 8178 trajectories,
comprising 69.03\% \texttt{Answer}, 16.53\% \texttt{ExitIntent},
10.28\% \texttt{Clarify}, and 4.16\% \texttt{Handoff}.

\subsection{Tool-Loop Recovery Augmentation}
Removing erroneous behavior does not teach the student how to recover from it. In particular, we observe that in approximately 5–7\% of cases, the student repeatedly invokes the same tool after receiving an empty result, even when an inference-time guard explicitly indicates that the invocation has already been performed.

To address this train–inference mismatch, we introduce synthetic tool-loop recovery examples. We identify trajectories in which the first tool invocation
returns an empty result and insert two additional identical calls. The third invocation is rejected with the exact feedback produced by the production guard. The inserted calls are masked from the SFT loss, so the student learns the recovery response without being trained to repeat the calls. We consider two recovery strategies:

\paragraph{Exit recovery (AugExit).}
The model is supervised to terminate the interaction
using \texttt{ExitIntent} after detecting the repeated
invocation.

\paragraph{Continuation recovery (AugLoop).}
The guard observation is followed by the original teacher
continuation, allowing the model to learn to proceed with
the task rather than repeat the blocked invocation.

This approach exposes the student to the same corrective feedback encountered at inference time while avoiding direct imitation of the undesired loop behavior.

\subsection{Training and Ablations}
Table~\ref{tab:sft-training} summarizes the training configuration for full-parameter SFT experiments.

All teacher-trajectory ablations share the same base model, training procedure, and evaluation protocol, differing only in the training dataset. We compare two augmentation
ratios (2\% and 5\%) for AugExit against AugLoop (5\%). To assess data efficiency, we additionally train AugExit (5\%) on a randomly sampled 50\% subset of the training data. Smaller subsets (10\% and 25\%) failed to maintain valid tool-call formatting and exhibited repeated tool-use loops; therefore, we exclude them from the quantitative comparison.

\begin{table}[t]
\centering
\small
\begin{tabular}{ll}
\toprule
Parameter & Value \\
\midrule
Policy initialization & Qwen3.6-27B \\
Max sequence length & 32768 tokens \\
Precision & bf16 \\
Optimizer & AdamW \\
Adam $(\beta_1,\beta_2)$ & $(0.9,0.98)$ \\
Weight decay & 0.1 \\
Gradient clipping (max norm) & 2.0 \\
Epochs & 4 \\
Global batch size & 128 \\
Peak learning rate & $2 \times 10^{-5}$ \\
Min learning rate & 0 \\
LR schedule & Cosine annealing \\
LR warmup & 7\% of schedule \\
\bottomrule
\end{tabular}
\caption{Teacher-trajectory SFT training configuration.}
\label{tab:sft-training}
\end{table}

\begin{table*}[t]
\centering
\small
\setlength{\tabcolsep}{5pt}
\begin{tabular}{@{}lrrrrrr@{}}
\toprule
Model / configuration
& Halluc.\,$\downarrow$
& Intent\,$\uparrow$
& Answer
& Clarify
& Handoff
& ExitIntent \\
\midrule
GPT-5.5
& 17.5 & 86.0 & 71.7 & 5.2 & 10.5 & 12.5 \\
\quad + self-critique
& 15.8 & 82.9 & 70.9 & 7.3 & 10.5 & 11.3 \\
\midrule
\textbf{AugExit (5\%) + hallucination filtering}
& 13.6 & 80.4 & 60.2 & 5.9 & 11.0 & 22.9 \\
AugExit (5\%)
& 19.5 & 78.3 & 70.5 & 3.2 & 11.1 & 15.2 \\
AugLoop (5\%)
& 23.4 & 79.1 & 68.5 & 3.5 & 11.2 & 16.9 \\
AugExit (2\%)
& 27.2 & 77.8 & 67.7 & 5.5 & 11.9 & 14.9 \\
AugExit (5\%, 50\% data)
& 27.4 & 73.7 & 51.9 & 13.0 & 14.3 & 20.8 \\
\midrule
SFT on rewritten operator replies
& 23.7 & 55.2 & 74.6 & 5.2 & 11.7 & 8.5 \\
\bottomrule
\end{tabular}
\caption{Offline evaluation of Qwen3.6-27B SFT variants
and GPT-5.5 on the referral-support benchmark.
Hallucination rate (Halluc.), intent match (Intent),
and outcome rates are percentages averaged over three runs,
with GPT-5.5 judging hallucinations and intent match.
AugExit and AugLoop denote exit-based and continuation-based
tool-loop recovery; percentages in configuration names indicate
augmentation ratios. Hallucination filtering uses GPT-5.5,
and 50\% data denotes training on half the dataset.
All student checkpoints are evaluated at epoch 2.
Bold marks the final SFT baseline.}
\label{tab:rf-v3-answer-types}
\end{table*}

Table~\ref{tab:rf-v3-answer-types} reports results on the
offline benchmark (App.~\ref{app:bench}), with all
student checkpoints evaluated at epoch 2. AugExit (5\%) achieves a lower hallucination rate than AugLoop (5\%) and AugExit (2\%), supporting the use of explicit exit recovery at the higher augmentation ratio. Halving the training data increases hallucinations and reduces the text-answer rate, indicating that the smaller dataset does not preserve the full-data policy's behavior.

Hallucination filtering further improves factual reliability, but shifts the policy toward fewer text answers and more exits. The AugExit (5\%) + Hallucination Filter checkpoint achieves the lowest hallucination rate among the evaluated SFT variants.

\subsection{Beyond Teacher Supervision}
\label{app:beyondSFT}
SFT on teacher-generated trajectories produces a capable tool-using policy, and filtering and recovery augmentation improve the agent's reliability. However, its supervision depends on the quality and
coverage of the teacher's demonstrations, which can
contain grounding errors. Filtering removes these
flawed examples without providing corrected alternatives. In our experiments, the resulting SFT policy improves factual reliability at the cost of lower answer coverage.

RL directly optimizes grounding, coverage, and response
requirements rather than relying solely on imitation.
SFT can reproduce facts from teacher demonstrations
without retrieving their supporting evidence.
GRPO directly penalizes such claims by assessing factual
precision against evidence available in the agent's
own trajectory (App.~\ref{app:grpo}). In the online research test for candidate-selection, assessors flagged unsupported
factual claims in 5.0\% of SFT answers, compared with
0.6\% for the final GRPO agent (App.~\ref{app:online}).

Neither approach eliminates grounding errors.
The following constructed examples illustrate two
distinct failure mechanisms. An SFT agent might state, ``Your bonus will arrive
within eight days,'' without a source supporting that
deadline. A GRPO agent might retrieve ``The status is
updated within 96 hours after all requirements are met,''
but answer, ``Your bonus will arrive within 96 hours.''
The first introduces an unsupported deadline, while
the second confuses a status update with a reward payment.

\section{Prompt Evolution Baseline}
\label{app:evolution}

We evaluated evolutionary optimization of the referral-support
agent's prompt without updating model weights,
testing whether instruction changes alone could provide
a reliable alternative to training for reducing hallucinations. We used ShinkaEvolve~\citep{lange2025shinka} within the
SkyDiscover framework~\citep{liu2026skydiscover}, with
GPT-5.5 proposing instruction changes and the off-the-shelf
Qwen3.6-27B serving as the support agent.

\subsection{Optimization Setup}

Candidates were Markdown bundles containing system instructions,
tool descriptions, and descriptions of structured output types and
fields. Optimization modified these textual components while retaining
the agent's tool implementations, tool arguments, and structured
output schemas.

We scored each candidate by averaging $F_1(p,c)$ across
the optimization dialogues, where $p$ measures factual
grounding and $c$ measures coverage of verified operator
claims (App.~\ref{app:grpo-what}).
Judges assessed precision against evidence retrieved
in the agent's own trajectory and coverage against
verified claims in the historical operator reply.
Evolved instructions and tool descriptions were not
treated as factual evidence.
This objective differs from the final GRPO reward
(App.~\ref{app:grpo-final}).

We used 100 dialogues drawn from the GRPO training set,
spanning 12 problem categories and sharing no dialogues
with the 500-dialogue offline benchmark
(App.~\ref{app:bench}).
The evaluator assessed completeness using reviewed
answer requirements and supporting evidence, which
were not supplied to the agent.

The search started from instructions manually tuned
by ML engineers for online performance.
We generated and evaluated 50 candidate instruction sets,
using one agent rollout per dialogue with a fixed seed.
The selected instructions were re-evaluated on the optimization
set using a different seed and then evaluated on the
offline benchmark.
We refer to the starting and selected instructions as
Initial and Evolved, respectively.

\subsection{Results and Implications}

The revisions clarified required steps and
bonus information, strengthened relevance and coverage
requirements, and added a checklist for pending or
missing bonuses and restrictions on unsupported inferences.
These changes affected the \texttt{get\_application\_details}
tool description, the \texttt{Answer} output-type
description, and the self-critique fields description.

\begin{table}[t]
\centering\small
\begin{tabular}{@{}lrrr@{}}
\toprule
Prompt & Halluc. $\downarrow$ & Intent $\uparrow$ & $F_1$ $\uparrow$\\
\midrule
Initial & 68.6 & 89.0 & 0.440 \\
Evolved & 58.6 & 84.8 & 0.481 \\
\bottomrule
\end{tabular}
\caption{Initial and evolved prompts.
Hallucination rate (Halluc.) and intent match (Intent)
are percentages on the offline benchmark.
$F_1$ is measured on the optimization set and reported on a 0--1 scale.} 
\label{tab:evolution-results}
\end{table}

The evolved instructions improved mean $F_1$ by 4.1 p.p.
over the initial instructions on the optimization set (Table~\ref{tab:evolution-results}).
On the offline benchmark, the evolved variant
reduced the hallucination rate by 10 p.p., but also
reduced intent match by 4.2 p.p. The improvement
in grounding therefore came at the cost of addressing
customer needs.

The evolved prompt still yielded a hallucination rate
of 58.6\%, compared with 8.6\% for the final GRPO agent.
In this setting, GRPO training was substantially more
effective than prompt optimization at reducing
hallucinations.

\section{Online Experiments}
\label{app:online}

Online evaluation compared alternative training strategies,
assessed the selected GRPO model against the original
procedural bot in a product A/B experiment, and monitored
its performance after deployment. Our primary online operational metrics are automation, the share of conversations closed without a human operator, and operator time per user, the average time human operators
spend handling a customer's conversations.

\subsection{Candidate Comparison}
\label{app:candcomparison}
A candidate-selection trial compared agent variants using
GPT-5.5, GRPO training, or SFT. We also included the style-reward GRPO variant to test whether operator-style imitation was a useful proxy
for task completion. Offline, this variant improved
intent match relative to the policy-reward variant,
but also produced more hallucinations (App.~\ref{app:grpo}).
Despite this tradeoff, we evaluated it online to determine
whether the apparent helpfulness gains translated into
higher automation.

The nine-day experiment included 19118 customers
across five arms, with 3778--3919 customers per arm.
Each arm was allocated 10\% of eligible traffic. Randomization was performed at the customer level among conversations invoking the
referral-support procedure.

The final GRPO candidate achieved the highest automation
among the trained agents, reaching 84.41\% compared
with 82.18\% for SFT and 63.3\% for production control.
It also reduced operator time per user by 28.6\%,
compared with 26.9\% for SFT
(Table~\ref{tab:online-training}). The style-reward variant achieved the lowest automation
among trained candidates (78.58\%) and reduced operator
time per user by only 11.0\%, cautioning against reward
selection based on operator-style similarity and
offline intent match alone.

\begin{table}[t]
\centering\small
\setlength{\tabcolsep}{3pt}
\begin{tabular}{@{}lccc@{}}
\toprule
Agent & Automation
      & \shortstack{$\Delta$ Auto.\\(p.p.)}
      & \shortstack{$\Delta$ Op. time\\per user} \\
\midrule
Procedural bot & 63.30\% & $0.00$ & $0.00\%$ \\
GPT-5.5      & 85.21\% & $+21.91$ & $-29.22\%$ \\
SFT          & 82.18\% & $+18.88$ & $-26.90\%$ \\
GRPO (style) & 78.58\% & $+15.28$ & $-11.00\%$ \\
\textbf{GRPO (final)} & 84.41\% & $+21.11$ & $-28.60\%$ \\
\bottomrule
\end{tabular}
\caption{Agent variants in the candidate-selection trial.
Changes are relative to production control,
in p.p. for automation and \% for operator time per user. The bold label marks the selected checkpoint.}
\label{tab:online-training}
\end{table}

Expert review assessed 800 conversations per model
to determine whether the trained candidates met
the consultation-quality requirement for deployment.
All three passed the established 5\% acceptance
threshold. The final GRPO candidate had the lowest
observed critical-error rate of 0.9\%
(95\% CI 0.4--1.8\%), compared with 3.1\% for SFT
(95\% CI 2.1--4.6\%) and 2.8\% (95\% CI 1.8--4.1\%) for style-reward GRPO.

Within the same review, assessors also flagged factual
claims unsupported by the KB passages and tool outputs
available in the agent's trajectory, regardless of
whether they constituted a critical consultation error.
Such claims appeared in 0.6\% of final GRPO answers,
compared with 5.0\% for SFT. 
These quality advantages, alongside the stronger
operational results, motivated the selection of
the final GRPO for product evaluation.

\subsection{Product A/B Evaluation}
The selected GRPO agent was evaluated against the
production control using customer-level randomization.
The procedural bot supported 29 intents organized into
20 predefined scenarios.
The design specified an eight-day experiment with
45\% of eligible traffic allocated to each arm.
The automation analysis included 17552 control
conversations and 17745 GRPO-agent conversations. Confidence intervals and significance tests for
operational metrics used bootstrap resampling
at the customer level within each arm, retaining
all conversations from each sampled customer.

\begin{table*}[t]
\centering\small
\setlength{\tabcolsep}{4pt}
\begin{tabular}{@{}lrrr@{}}
\toprule
Metric & Procedural bot & GRPO (final) & $\Delta$ [95\% CI] \\
\midrule
Automation
& 63.3\% & 80.8\% & $+17.5$ [$16.4$, $18.8$] p.p. \\
Operator time/conversation (s)
& 109.671 & 72.231 & $-34.14\%$ \\
Operator time/user (s)
& 134.058 & 93.304 & $-30.40\%$ [$-34.3\%$, $-27.1\%$] \\
Messages/conversation
& 15.074 & 7.855 & $-47.89\%$ \\
Mean CSAT
& 4.451 & 4.376 & $-0.075$ \\
Negative CSAT
& 18.8\% & 20.4\% & $+1.6$ p.p. \\
CSAT response rate
& 10.1\% & 8.8\% & $-1.3$ p.p. \\
Technical-failure handoffs
& 1.7\% & 0.6\% & $-1.1$ p.p. \\
Repeat contact within 24 hours
& 25.6\% & 21.9\% & $-3.7$ p.p. \\
Repeat contact within 48 hours
& 34.1\% & 28.9\% & $-5.2$ p.p. \\
Repeat contact within five days
& 42.7\% & 39.8\% & $-2.9$ p.p. \\
\bottomrule
\end{tabular}
\caption{Product A/B results comparing the procedural
bot with GRPO (final). Changes are relative to the
procedural bot, with units shown in the $\Delta$ column. Brackets report 95\% customer-level bootstrap
confidence intervals for the 2 primary
operational effects.}
\label{tab:online-original-bot}
\end{table*}

\begin{table*}[t]
\centering\small
\setlength{\tabcolsep}{4pt}
\begin{tabular}{@{}lrrrrr@{}}
\toprule
Error category & Procedural bot & GRPO (final) & $\Delta$ (p.p.) & $p$ \\
\midrule
At least one error & 25.9\% & 9.3\% & $-16.5$ & $<0.001$ \\
Failure to address negative sentiment & 2.1\% & 0.7\% & $-1.3$ & 0.1990 \\
Unnecessary question & 0.0\% & 0.0\% & 0.0 & --- \\
Advice unrelated to customer needs & 4.4\% & 6.6\% & $+2.2$ & 0.2052 \\
Question unrelated to customer needs & 7.6\% & 3.2\% & $-4.5$ & 0.0078 \\
Action unrelated to customer needs & 17.1\% & 1.5\% & $-15.6$ & $<0.001$ \\
\bottomrule
\end{tabular}
\caption{Expert-assessed consultation errors in the
product A/B test. Differences
are relative to the procedural bot.}
\label{tab:online-consultation-errors}
\end{table*}

Table~\ref{tab:online-original-bot} compares the original
bot with the selected final GRPO agent.
Automation increased from 63.3\% to 80.8\%,
a difference of 17.5 percentage points.
Operator time per user decreased by 30.40\%,
operator time per conversation was 34.14\% lower,
and conversations contained 47.89\% fewer messages.

The agent has one tool the procedural bot lacked,
\texttt{find\_missing\_application}.
It was called in 2.0\% of agent-arm conversations.
Even if every such conversation was automated because
of the tool, its direct contribution would account
for at most 2 percentage points of the observed
automation gain.

Technical-failure handoffs declined from 1.7\% to 0.6\%.
Repeat-contact rates decreased from 25.6\% to 21.9\%
within 24 hours, from 34.1\% to 28.9\% within 48 hours,
and from 42.7\% to 39.8\% within five days.
Customer requests for an operator decreased from
11.8\% to 11.7\% of conversations.

Mean CSAT was slightly lower (4.38 vs.\ 4.45)
and negative CSAT slightly more frequent
(20.4\% vs.\ 18.8\%). Neither difference was
statistically significant.

\subsection{Expert Assessment of Consultation Quality}
\label{app:post-deploy}
Domain experts assessed whether the agent's responses and actions addressed customer needs. The share of evaluated interactions containing at least one consultation error decreased from 25.9\% in the control to 9.3\% with the GRPO agent, a reported relative reduction of 63.9\% ($p<0.001$). This comparison included 747 evaluated interactions.

The largest improvement concerned actions unrelated to customer needs. Questions unrelated to those needs also became less frequent. Advice unrelated to customer needs increased numerically, but the difference was not statistically significant (Table~\ref{tab:online-consultation-errors}).

A separate domain-specific assessment covered 722 segments from the GRPO arm. The critical-error rate was 0.7\% (95\% CI 0.3–1.6\%, n = 722), which satisfies the 5\% acceptance threshold.

\subsection{Post-Deployment Monitoring}

Post-deployment monitoring covered approximately five weeks, including four complete calendar weeks. Across these weeks, mean daily automation ranged from 84.4\% to 87.8\%, compared with 80.8\% in the product A/B evaluation. Mean daily CSAT remained between 4.08 and 4.10 (Table~\ref{tab:online-monitoring}). The agent operates in an asynchronous support chat.
End-to-end latency includes queueing, model generation,
and tool execution, measured from receipt of the
customer's message to completion of the response.
Daily p95 latency remained within the predefined
90-second serving budget throughout monitoring. No operational incidents were recorded during the monitoring period.

\begin{table*}[!t]
\centering
\small
\begin{tabular}{@{}lrr@{}}
\toprule
 & \multicolumn{2}{c}{Mean daily value} \\
\cmidrule(l){2-3}
Period & Automation & CSAT \\
\midrule
Sep.\ 21--27 & 84.4\% & 4.09 \\
Sep.\ 28--Oct.\ 4 & 84.9\% & 4.10 \\
Oct.\ 5--11 & 87.8\% & 4.08 \\
Oct.\ 12--18 & 85.8\% & 4.10 \\
\bottomrule
\end{tabular}
\caption{Post-deployment monitoring over four complete weeks in 2026. Values are unweighted means of daily metrics.}
\label{tab:online-monitoring}
\end{table*}

Expert review continued after deployment. Across 19 days with available review results within the first 25 calendar days, consultation errors were recorded in 1644 of 13615 reviewed consultations (12.07\%). This measure covers consultation errors generally and should not be interpreted as a critical-error rate.

These observations show sustained automation and stable weekly CSAT during the monitoring period. Comparisons with the A/B results are descriptive because traffic composition and review samples may differ between periods.

\subsection{Human Quality Assessment Protocol}
\label{app:annotation}
Production conversations are assessed by a dedicated group of internal professional assessors. Assessors complete domain-specific training and qualify through an examination using gold-annotated examples before contributing to the quality assessment. Review samples were drawn randomly from each arm
over the same period, without selection by outcome
or suspected error. All arms used the same inclusion
criteria and segment definitions. Cases from different
arms were mixed, and assessors were blinded to model
identity, experimental arm, and automated judge scores.
They received the same types of conversational context
and customer records and applied a predefined
critical-error rubric. Assessors reviewed the agent's customer-facing
consultation and marked errors under the domain
rubric, including whether an error was critical.

Assessors assign error categories. A category is included in the final annotation when at least three of five assessors select the same category. The quality team reported that approximately 95\% of reviewed items reached this three-of-five category consensus.

Approximately 700 production interaction segments were reviewed after the A/B test. The critical-error rate is the share of reviewed segments with at least one consensus critical error, assessed against a 5\% acceptance threshold.

\end{document}